\documentclass[12pt]{colt2023} 

\title[Radial Neural Networks]{Universal approximation and model compression for radial neural networks}
\usepackage{times}

\coltauthor{%
 \Name{Iordan Ganev}  \Email{iganev@cs.ru.nl}\\ 
 \addr  Radboud University
 \AND
 \Name{Twan van Laarhoven}  \Email{tvanlaarhoven@cs.ru.nl}\\
 \addr  Radboud University
 \AND
 \Name{Robin Walters}  	  	\Email{r.walters@northeastern.edu} \\
 	\addr  	Northeastern University 
}

\usepackage{amsmath,amsfonts,bm}
\usepackage{mathrsfs} 
\usepackage{comment}
\usepackage{tikz-cd}

\usepackage{tikz}
\usetikzlibrary{arrows,chains,matrix,positioning,scopes}
\usetikzlibrary{matrix}
\usetikzlibrary{arrows.meta}
\tikzset{|/.tip={Bar[width=.8ex,round]}}
\usepackage{pgfplots}
\pgfplotsset{compat=1.16}

\usepackage{thmtools,thm-restate}
\usepackage[ruled]{algorithm2e}
\usepackage[all,2cell]{xy} \UseAllTwocells \SilentMatrices \SelectTips{cm}{}

\newcommand{\bfA}{\mathbf{A}}
\newcommand{\bfB}{\mathbf{B}}
\newcommand{\bfQ}{\mathbf{Q}}
\newcommand{\bfR}{\mathbf{R}}
\newcommand{\bfT}{\mathbf{T}}
\newcommand{\bfU}{\mathbf{U}}
\newcommand{\bfV}{\mathbf{V}}
\newcommand{\bfW}{\mathbf{W}}
\newcommand{\hbfW}{\hat{\bfW}}
\newcommand{\bfWred}{\mathbf{W}^\text{\rm red}}
\newcommand{\hbfWred}{\hat{\bfW}^\text{\rm red}}

\newcommand{\bfX}{\mathbf{X}}

\newcommand{\bfZ}{\mathbf{Z}}

\newcommand{\boldrho}{\boldsymbol \rho}

\newcommand{\bfb}{\mathbf{b}}
\newcommand{\bfbred}{\mathbf{b}^\text{\rm red}}

\newcommand{\bfc}{\mathbf{c}}

\newcommand{\bfg}{\mathbf{g}}
\newcommand{\bfh}{\mathbf{h}}

\newcommand{\bfn}{\mathbf{n}}

\newcommand{\bft}{\mathbf{t}}
\newcommand{\bfnred}{\mathbf{n}^{\text{\rm red}}}

\newcommand{\nred}{n^{\rm red}}

\newcommand{\Fred}{F^{\rm red}}
\newcommand{\rhored}{\rho^{\rm red}}
\newcommand{\brhored}{\boldrho^{\rm red}}

\newcommand{\inv}{^{-1}}

\newcommand{\id}{\text{\rm id}}
\newcommand{\inx}{i}

\newcommand{\inc}{\text{\rm inc}}
\newcommand{\Inc}{\text{\rm Inc}}
\newcommand{\ext}{\text{\rm ext}}

\newcommand{\R}{\mathbb{R}}

\renewcommand{\L}{\mathcal{L}}

\newcommand{\Hom}{\mathrm{Hom}}

\newcommand{\GL}{\mathrm{GL}}

\newcommand{\Par}{\mathsf{Param}}
\newcommand{\Parn}{\Par( \bfn)}
\newcommand{\Parnred}{\Par( \bfnred)}
\newcommand{\Parintn}{\Par^{\text{\rm int}} ( \bfn)}

\newcommand{\Proj}{\mathrm{Proj}}

\newcounter{prooftheorem}
\newtheorem{sublemma}{Lemma}[prooftheorem]
\newtheorem{subprop}[sublemma]{Proposition}
\newcommand{\widthsublemmas}[1][\thetheorem]{%
	\setcounter{prooftheorem}{#1}%
	\setcounter{sublemma}{0}%
}
\newcommand{\widthsublemmasref}[1]{%
	\setcounterref{prooftheorem}{#1}%
	\setcounter{sublemma}{0}%
}

\usepackage{amsmath,amsfonts,bm}

\def\eqref#1{equation~\ref{#1}}

\def\1{\bm{1}}

\DeclareMathAlphabet{\mathsfit}{\encodingdefault}{\sfdefault}{m}{sl}
\SetMathAlphabet{\mathsfit}{bold}{\encodingdefault}{\sfdefault}{bx}{n}

\begin{document}
	
	\pagestyle{plain}

\maketitle

\begin{abstract}%
	We introduce a class of fully-connected neural networks whose activation functions, rather than being pointwise, rescale feature vectors by a function depending only on their norm. We call such networks {\it radial} neural networks, extending previous work on rotation equivariant networks that considers rescaling activations in less generality. 
We prove universal approximation theorems for radial neural networks, including in the more difficult cases of bounded widths and unbounded domains. Our proof techniques are novel, distinct from those in the pointwise case.
Additionally, radial neural networks exhibit a rich group of orthogonal change-of-basis  symmetries on the vector space of trainable parameters. Factoring out these symmetries leads to a practical  lossless model compression algorithm.  Optimization of the compressed model by gradient descent is equivalent to projected gradient descent for the full model.
\end{abstract}


\section{Introduction}\label{sec:intro}

Inspired by biological neural networks, the theory of artificial neural networks has largely focused on pointwise (or ``local'') nonlinear layers \citep{rosenblatt1958perceptron, cybenko1989approximation}, in which the same function $\sigma \colon \mathbb{R} \to \mathbb{R}$ is applied to each coordinate independently:
\begin{equation}\label{eq:ptwise-act} 
	\R^n \to \R^n, \qquad
	v = (v_1 \ , \ \dots \ , \ v_n) \  \mapsto \  (\sigma(v_1) \ , \ \sigma(v_2) \ , \ \dots \ , \ \sigma(v_n)).
\end{equation}   
In networks with pointwise nonlinearities, the standard basis vectors in $\R^n$  can be interpreted as ``neurons'' and the nonlinearity as a ``neuron activation.''  
Research has generally focused on finding functions $\sigma$ which lead to more stable training,
have less sensitivity to initialization, or are better adapted to certain applications \citep{ramachandran2017searching, 
	misra2019mish, milletari2018mean, clevert2015fast, klambauer2017self}.   
Many $\sigma$ have been considered, including sigmoid, ReLU, arctangent, ELU, Swish, and others.

However, by setting aside the biological metaphor, it is possible to consider a much broader class of nonlinearities, which are not necessarily pointwise, but instead depend simultaneously on many coordinates. {Freedom from the pointwise assumption allows one to design activations that yield expressive function classes with specific advantages. Additionally, certain choices of non-pointwise activations maximize symmetry in the parameter space of the network, leading to compressibility and other desirable properties.}

In this paper, we introduce \emph{radial} neural networks which employ non-pointwise nonlinearities called \emph{radial rescaling} activations. 
Such networks enjoy several provable properties, including high model compressibility, symmetry in optimization, and universal approximation. 
Radial rescaling activations are defined by rescaling each vector by a scalar  that depends  only on the norm of the vector:
\begin{equation}\label{eq:radial-intro}
	\rho: \R^n \to \R^n, \qquad v  \  \mapsto \ \lambda(|v|) v, 
\end{equation} 
where $\lambda$ is a scalar-valued function of the norm. Whereas in the pointwise setting, only the linear layers mix information between different components of the latent features, for radial rescaling activations, all coordinates of the activation output vector are affected by all coordinates of the activation input vector.   
The inherent geometric symmetry of radial rescalings makes them particularly useful for designing equivariant neural networks \citep{weiler_general_2019, 	sabour2017dynamic, weiler20183d, weiler2018learning}. 

We note that radial neural networks constitute a simple and previously unconsidered type of multilayer radial basis functions network  \citep{broomhead1988radial}, namely, one where the number of hidden activation neurons (often denoted $N$) in each layer is equal to one. Indeed, pre-composing equation \ref{eq:radial-intro} with a translation and post-composing with a linear map, one obtains a special case of the local linear model extension of a radial basis functions network.

\begin{figure}[t]
	\centering
	\includegraphics[width=\textwidth]{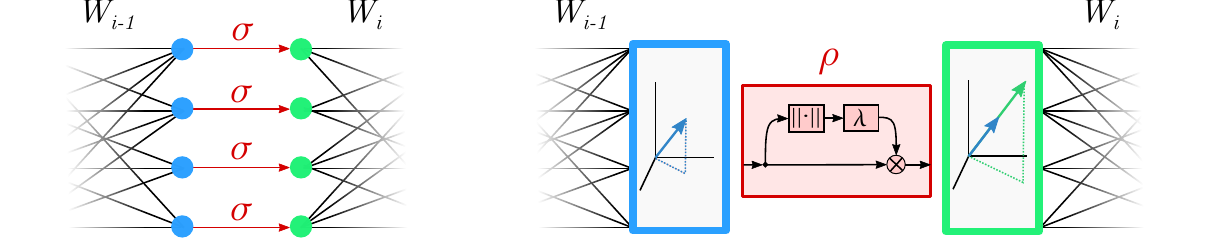}
	\caption{(Left) Pointwise activations distinguish a specific basis of each hidden layer  and treat each coordinate independently, see \eqref{eq:ptwise-act}. 
		(Right) Radial rescaling activations rescale each feature vector by a function of the norm, see \eqref{eq:radial-intro}.
	}
	\label{fig:ptwise-vs-radial}
\end{figure}

In our first set of main results, we prove that radial neural networks are in fact {\it universal approximators}.
Specifically, we demonstrate that  any {asymptotically affine} function can be approximated with a radial neural network, suggesting potentially good extrapolation behavior. Moreover, this approximation can be done with {bounded width}.
Our approach to proving these results departs markedly from techniques used in the pointwise case. 
{Additionally, our result is not implied by the universality property of radial basis functions networks in general, and differs in significant ways, particularly in the bounded width property and the approximation of asymptotically affine functions.}

In our second set of main results, we exploit parameter space symmetries of radial neural networks to achieve {\it model compression}. Using the fact that radial rescaling activations commute with orthogonal transformations, we develop a practical algorithm to systematically factor out orthogonal symmetries via iterated QR decompositions. This leads to another radial neural network with fewer neurons in each hidden layer. The resulting model compression algorithm is {\it lossless}: the compressed network and the original network both have the same value of the loss function on any batch of training data. 


Furthermore, we prove that the loss of the compressed model after one step of gradient descent is equal to the  loss of the original model after one  step of \emph{projected gradient descent}. As explained below, projected gradient descent involves zeroing out certain parameter values after each step of gradient descent. Although training the original network  may result in a lower loss function after fewer epochs, in many cases the compressed network takes less time per epoch to train and is faster in reaching  a local minimum.

To summarize, our main contributions and headline results are:

\begin{itemize}
	\item \emph{Radial rescaling activations are an alternative to pointwise activations:} We provide a formalization of radial neural networks, a new class of neural networks;
	\item \emph{Radial neural networks are universal approximators:} Results include a) approximation of asymptotically affine functions, and b) bounded width approximation;
	\item \emph{Radial neural networks are inherently compressible:} We prove a lossless compression algorithm for such  networks and a theorem providing the relationship between optimization of the original and compressed networks.
	\item \emph{Radial neural networks have practical advantages:} We describe experiments verifying all theoretical results and showing that radial networks outperform pointwise networks on a noisy image recovery task.
\end{itemize}

\section{Related work}\label{sec:related-work}

{\bf Radial rescaling activations.}
{As noted, radial rescaling activations are a special case of the activations used in radial basis functions networks \citep{broomhead1988radial}.}
Radial rescaling functions have the symmetry property of preserving vector directions, and hence exhibit rotation equivariance.  Consequently, examples of such functions, such as the squashing nonlinearity and Norm-ReLU,  feature in the study of  rotationally equivariant neural networks   \citep{weiler_general_2019, sabour2017dynamic, weiler20183d, weiler2018learning, jeffreys_kahler_2021}.   
However, previous works apply the activation only along the channel dimension, and consider the orthogonal group $O(n)$ only for $n=2,3$.  In contrast,  we apply the activation across the entire hidden layer, and $O(n)$-equivariance where $n$ is the hidden layer dimension. 
Our constructions echo the  vector neurons formalism \citep{deng_vector_2021}, in which the output of a nonlinearity is a vector rather than a scalar.

{\bf Universal approximation.}
Neural networks of arbitrary width and sigmoid activations have long been known to be universal approximators  \citep{cybenko1989approximation}. Universality can also be achieved by bounded width networks with arbitrary depth \citep{lu2017expressive}, and generalizes to other activations and architectures \citep{hornik1991approximation,  yarotsky2022universal, ravanbakhsh2020universal, sonoda2017neural}. Previous work has also considered the interaction between learnability, approximation capacity, and depth separation; these are related to open problems in computational complexity \citep{malach_connection_2021, vardi_size_2021}.  While most work has focused on compact domains, some recent work also considers non-compact domains \citep{kidger2020universal, wang2022approximation}, {but only for $L_p$ functions, which are less general than asymptotically affine functions}.
The techniques used for pointwise activations  do not generalize to radial rescaling activations, where all activation output coordinates are affected by all input coordinates. Consequently, individual radial neural network approximators of two different functions cannot be easily combined to an approximator of the sum of the functions. {The standard proof of universal approximation for radial basis functions networks requires an unbounded increase the number of hidden activation neurons, and hence does not apply to the case of radial neural networks \citep{park1991universal}.}

{\bf Groups and symmetry.} 
Appearances of symmetry in machine learning have generally focused on symmetric input and output spaces. Most prominently, equivariant neural networks  incorporate symmetry as an inductive bias and feature weight-sharing constraints based on equivariance. Examples include   $G$-convolution, steerable CNN, and Clebsch-Gordon networks \citep{cohen2019gauge,  weiler_general_2019, cohen2016group, chidester2018rotation, kondor_generalization_2018, bao2019equivariant, worrall2017harmonic, cohen2016steerable, weiler2018learning, dieleman2016cyclic, lang2020wigner, ravanbakhsh2017equivariance}. 
By contrast, our approach  does not depend on symmetries  of the input domain, output space, or feedforward mapping.  Instead, we exploit parameter space symmetries and  obtain  results that apply to domains with no apparent symmetry.

{\bf Model compression.}
A major goal in machine learning is to find methods to reduce the number of trainable parameters, decrease memory usage, or accelerate inference and training \citep{cheng2017survey, zhang2018systematic}.  Our approach toward this goal  differs significantly from most existing methods in that it is based on the inherent symmetry of  network parameter spaces.  One prior method is \emph{weight pruning}, which removes redundant  weights with little loss in accuracy \citep{han2015deep, blalock2020state, karnin1990simple}.  Pruning can be done during training 
\citep{frankle2018lottery} or at initialization \citep{	lee2019signal, wang2020picking}.  \emph{Gradient-based pruning} removes weights by estimating the increase in loss resulting from their removal \citep{	lecun1990optimal, hassibi1993second, dong2017learning, molchanov2016pruning}.    A complementary approach is \emph{quantization}, which decreases the bit depth of weights \citep{wu2016quantized, howard2017mobilenets, gong2014compressing}.    \textit{Knowledge distillation} identifies a small model mimicking the performance of a larger model  \citep{bucilua2006model,hinton2015distilling, ba2013deep}. 
\textit{Matrix Factorization} methods replace fully connected layers with lower rank or sparse factored tensors \citep{cheng2015fast, cheng2015exploration, tai2015convolutional, lebedev2014speeding, rigamonti2013learning, lu2017fully} and can often be applied before training.   Our method  involves a type of matrix factorization based on the QR decomposition; however, rather than aim for rank reduction, we leverage this decomposition to reduce hidden widths via change-of-basis operations on the hidden representations.  Close to our method are lossless compression methods which remove stable neurons in ReLU networks \citep{	serra2021scaling, serra2020lossless} or  exploit permutation parameter space symmetry to remove neurons  \citep{sourek2020lossless}; our compression instead follows from the symmetries of the radial rescaling activation. Finally, the compression results of \cite{jeffreys_kahler_2021}, while conceptually similar to ours, are weaker, as the unitary group action is only on disjoint layers, and the results are only stated for   the squashing nonlinearity.

\section{Radial neural networks}\label{sec:rad-nns}

In this section, we define radial rescaling functions and radial neural networks. Let $h : \R \to \R$ be a function.  For any $n \geq 1$, set:
\[ h^{(n)} : \R^n \to \R^n \qquad \qquad
h^{(n)}(v)  =  h(\lvert v \rvert) \frac{v}{|v|} \]
for $v \neq 0$, and $h^{(n)}(0) = 0$. A function $\rho : \R^n \to \R^n$ is called a {\it radial rescaling} function if $\rho = h^{(n)}$ for some piecewise differentiable $h : \R \to \R$. Hence, $\rho$ sends each input vector to a scalar multiple of itself, and that scalar depends only on the norm of the vector\footnote{A function $ \R^n \to \R$ that depends only on the norm of a vector is known as a {\it radial} function. Radial rescaling functions rescale each vector according to the radial function $v \mapsto \lambda(|v|) := \frac{h(|v|)}{|v|}$. This explains the connection to Equation \ref{eq:radial-intro}.}. It is easy to show that radial rescaling functions commute with orthogonal transformations.

\begin{figure}
	\centering
	\begin{tikzpicture}
	\def\xmax{2.1}
	\begin{axis}[
    	xmin=-\xmax,xmax=\xmax, 
    	ymin=-2.2, ymax=2.2, samples=71,
    	clip=false,
    	width=0.8\textwidth,
    	height=0.45\textwidth,
    	axis lines=center,
    	xtick={-3,-2.5,...,3},
    	ytick={-5,-4.5,...,5},
    	legend style={draw=none,fill=none},
    	legend cell align={left},
    	every axis legend/.append style={at={(0.52,1)}, anchor=north west},
    	line4/.style={thick, color={rgb,10:green,3;blue,8}},
    	line3/.style={thick, color={rgb,11:red,10;green,7}},
    	line2/.style={thick, color={rgb,10:red,4;green,8}},
	]
	\addplot[line4, domain=-1:1] { 0 };
	\draw[dotted] (axis cs:-1,0) -- (axis cs:-1,-1);
	\draw[dotted] (axis cs:1,0) -- (axis cs:1,1);
	\addplot[line4,forget plot,domain=-\xmax:-1] { (x) };
	\addplot[line4,forget plot,fill=white,only marks,mark=*] coordinates{(-1,0)(1,0)};
	\addplot[line4,forget plot,only marks,mark=*] coordinates{(-1,-1)(1,1)};
	\addplot[line4,forget plot,domain=1:\xmax] { (x) }
	node[pos=1,black,right] {(1) $\text{Step-ReLU}(r)$};
	\addplot[line2,domain=-\xmax:\xmax] { max(0, abs(x) - 1/2)*sign(x) }
	node[pos=1,black,right] {(3) Shifted ReLU};
	\addplot[line3,domain=-\xmax:\xmax] { x^2/(x^2+1)*sign(x) }
	node[pos=1,black,right] {(2) Squashing function};
	\end{axis}
\end{tikzpicture}
	\caption{{Examples of different radial rescaling functions in $\R^1$, see Example \ref{ex:radialact}.}}
	\label{fig:radial_act}
\end{figure}
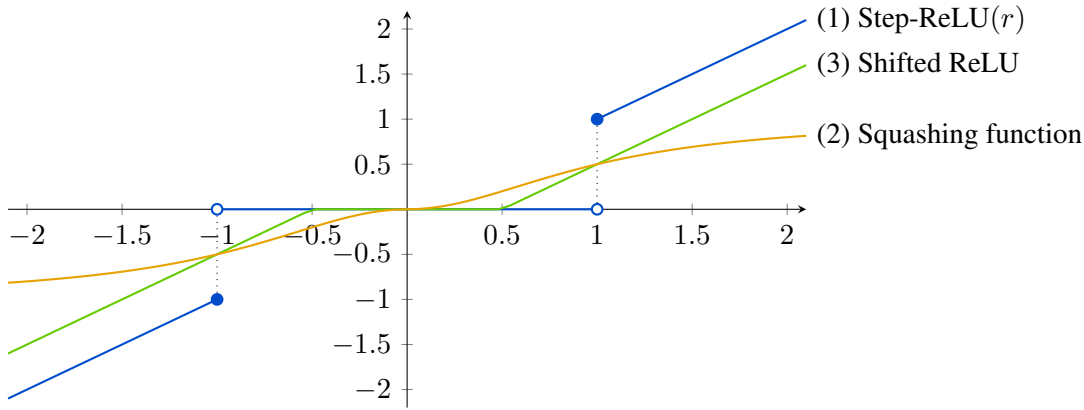

\begin{example}\label{ex:radialact} (1) Step-ReLU, where $h(r) = r$ if $r \geq 1$ and $0$ otherwise.  In this case, the radial rescaling function is given by
	\begin{equation}\label{eqn:step-relu}
		\rho: 	\R^n \rightarrow \R^n, \qquad
		v \mapsto v  \  \text{\rm if $|v| \geq 1$}; \qquad v \mapsto 0 \   \text{\rm if $|v| <1$}
	\end{equation}
	(2) The squashing function, where
	$h(r) = r^2/(r^2 + 1)$.
	(3) Shifted ReLU, where  $ h(r) = \max(0, r - b )$ for $r >0$ and $b$ is a real number.  See Figure \ref{fig:radial_act}. We refer to \cite{weiler_general_2019} and the references therein for more examples and discussion of radial functions. \end{example}


A {\it radial neural network} with $L$ layers consists of positive integers $n_i$ indicating the width of each layer $i = 0, 1, \dots, L$; the trainable parameters, comprising of a matrix $W_i \in \R^{n_i \times n_{i-1}}$ of weights and a bias vector $b_i \in \R^{n_i}$ for each $i = 1, \dots, L$; and a radial rescaling function 	$\rho_i :\R^{n_i} \to \R^{n_i}$ for each  $i = 1, \dots, L$.
We refer to the tuple $\bfn = (n_0, n_1, \dots, n_L)$ as the {\it widths vector} of the neural network. The hidden widths vector is $\bfn^\text{\rm hid} = (n_1, n_2, \dots, n_{L-1})$. The feedforward function $F : \R^{n_0} \to \R^{n_L}$ of a radial neural network  is defined in the usual way as an iterated composition of affine maps and activations. Explicitly, set $F_0 = \id_{\R^{n_0}}$ and the partial feedforward functions are:
\begin{equation*}
F_i : \R^{n_0} \to \R^{n_i}, \qquad   x\mapsto  \rho_i\left( W_i \circ F_{i-1}(x) + b_i\right)
\end{equation*}
for $i = 1, \dots, L$. Then the feedforward function is $F = F_L$.  {Radial neural networks are a special type of radial basis functions network; we explain the connection in Appendix \ref{app:rbfns}.}

\begin{remark}
	If $b_i = 0$ for all $i$, then we have $F(x) = W \left( \mu(x) x\right)$ where $\mu : \R^n \to \R$ is a scalar-valued function and $W = 	 W_L  W_{L-1} \cdots W_1 \in \R^{n_L \times n_0}$ is the product of the weight matrices. 
	If any of the biases are non-zero, then the feedforward function lacks such a simple form.
\end{remark}


\section{Universal Approximation}\label{sec:UA}

We now consider two universal approximation results. The first 
approximates asymptotically affine functions with a network of unbounded width. The second generalizes to bounded width.  Proofs appear in Appendix~\ref{app:UA}. Throughout, $B_r(c) = \{ x  \in \R^n  :  |x-c| <r \}$ is the $r$-ball around a point $c$, and an affine map $\R^n \to \R^m$ is one of the from $L(x)= Ax + b$ for  $A \in \R^{m \times n}$ and $b\in \R^m$.

\subsection{Approximation of asymptotically affine functions}\label{subsec:asymp-lin-approx}

A continuous function  $f : \R^n \to \R^m$ is  {\it asymptotically affine} if there exists an affine map $L : \R^n \to \R^m$ such that, for every $\epsilon >0$, there is a compact subset $K$ of $\R^n$ such that $ |L(x) - f(x)| < \epsilon$ for all $x \in \R^n \setminus K$. 
In particular, continuous functions with compact support are asymptotically affine.
The continuity of $f$ and compactness of $K$ imply that, for any $\epsilon >0$,  there exist $c_1, \dots, c_N \in K$ and $r_1, \dots, r_N \in (0,1)$ such that, first, the union of the balls $B_{r_i}(c_i)$ covers $K$ and, second, for all $i$, we have $f\left(B_{r_i} (c_i) \cap K \right) \subseteq  B_{\epsilon}(f(c_i))$. Let $N(f,K, \epsilon)$ be the minimal\footnote{{In many cases, the constant $N(f,K,\epsilon)$ can be bounded explicitly.} For example,  if $K$ is the unit cube in $\R^n$ and $f$ is Lipschitz continuous with Lipschitz constant $R$, then $N(f,K,\epsilon) \leq \Bigl\lceil\frac{R \sqrt{n}}{2\epsilon}\Bigr\rceil^n$.} choice of $N$.

\begin{figure}
	\centering
	\newbool{drawnodes}
\definecolor{spherecolor}{rgb}{0.1,0.8,0.1}
\definecolor{ambientcolor}{rgb}{0,0.2,1}
\definecolor{blobcolor}{rgb}{0.2,0,1}
\definecolor{spheremapcolor}{rgb}{0.9,0.1,0}
\begin{tikzpicture}
    [z={(0.5cm,0.5cm)},
     point/.style={circle,fill=black,inner sep=1pt},
     ball/.style={draw=black!80!white},
     blob/.style={fill=blue!25!white},
     ambient/.style={fill=blue!5!white, draw=blue!30},
    ]
    \newenvironment{projectxz}[1][90]{\begin{scope}[cm={1,0,sin(#1)*0.5,cos(#1)*1 + sin(#1)*0.5,(0,0)}]}{\end{scope}}
    \newenvironment{projectyz}[1][90]{\begin{scope}[cm={cos(#1)*1 + sin(#1)*0.5,sin(#1)*0.5,0,1,(0,0)}]}{\end{scope}}
    \newcommand{\ball}[4][]{
        \draw[ball,#1] #2 circle (#3cm);
        \ifbool{drawnodes}{
            \node[point,label={below:$#4$}] at #2 {};
        }{}
    }
    \newcommand{\blob}[2][]{
        \begin{scope}[x=0.9cm]
            \draw[draw=none,blob,even odd rule,#1] (0.5,0)
              .. controls +(1,0) and +(-1,0.5) .. (3,-0.5)
              .. controls +(1,-0.5) and +(-0.5,-0.5) .. (5,1)
              .. controls +(1,1) and +(0.5,-0.1) .. (4,3)
              .. controls +(-1,0.2) and +(0.5,0) .. (1,2.5)
              .. controls +(-2,0) and +(-0.5,0) .. cycle
              #2;
            \ifbool{drawnodes}{
                \node[color=blobcolor!80!black] at (5.0,0.3) {$K$};
            }{}
        \end{scope}
    }
    \newcommand{\cover}[1][]{
        \begin{scope}
            \ball[#1]{(0.4,1.9)}{0.9}{c_1};
            \ball[#1]{(2.3,1.3)}{0.9}{c_2};
            \ball[#1]{(1.7,2.55)}{0.9}{c_3};
            \ball[#1]{(0.9,0.7)}{1.2}{c_4};
            \ball[#1]{(2.3,0)}{0.7}{c_5};
            \ball[#1]{(3.4,0.1)}{0.9}{c_6};
            \ball[#1]{(3.3,2.5)}{1.0}{c_7};
            \ball[#1]{(3.9,1.6)}{1.1}{c_8};
        \end{scope}
    }
    \newcommand{\inputspace}{
        \draw[ambient] (-0.8,-1.2) rectangle (5.4,3.8);
        \ifbool{drawnodes}{
            \node[color=ambientcolor!80!black!80] at (-0.4,-0.9) {$\R^n$};
        }{}
    }
    \newcommand{\axes}{
        \draw[->,thick,color=blue] (-0.7,-1) -- (1.3,-1);
        \draw[->,thick,color=blue] (-0.7,-1) -- (-0.7,1);
    }
    \newcommand{\sphere}[2]{
        \begin{scope}[cm={1.1124,0.1124, 0.1124,1.1124, (0,0)}]
            \draw[spherecolor] (0,0) circle (#1);
        \end{scope}
        \begin{projectxz}\draw[spherecolor] (0,0) circle (#1);\end{projectxz}
        \foreach \a in {0,...,7} {
            \begin{projectyz}[\a*180/8+15]
                \draw[spherecolor] (#2-90:#1) arc[radius=#1, start angle=#2-90, end angle=270-#2];
                \draw[blue!5!white!70!spherecolor] (270-#2:#1) arc[radius=#1, end angle=-90+#2, start angle=-90-#2];
            \end{projectyz}
        }
    }
    \newcommand{\altsphere}[2]{
        \begin{scope}[cm={1.1124,0.1124, 0.1124,1.1124, (0,0)}]
            \draw[thick,spherecolor] (0,0) circle (#1);
            \foreach \a in {15,45,...,359} {
                \draw[spheremapcolor,->,thick] (\a:1*#1) -- (\a:0.4*#1);
            }
        \end{scope}
        \begin{projectxz}
            \draw[spherecolor] (0,0) circle (#1);
        \end{projectxz}
        \foreach \a in {1,...,8} {
            \begin{projectyz}[\a*180/8+15]
                \draw[spherecolor] (#2-90:#1) arc[radius=#1, start angle=#2-90, end angle=270-#2];
                \draw[blue!5!white!70!spherecolor] (270-#2:#1) arc[radius=#1, end angle=-90+#2, start angle=-90-#2];
            \end{projectyz}
        }
        \begin{scope}[cm={1.1124,0.1124, 0.1124,1.1124, (0,0)}]
        \end{scope}
    }
    
    \begin{scope}[scale=0.78]
        \begin{scope}
            \booltrue{drawnodes}
            \inputspace;
            \blob[]{}
            \cover;
        \end{scope}
        
        \begin{scope}[shift={(7.25,-0.6)}]
            \begin{projectxz}
                \inputspace;
                \blob[]{}
                \boolfalse{drawnodes}
                \cover[black!50!white];
                \draw[ball,fill=blue!50!spherecolor!5!white,draw=spherecolor] (2.3,1.3) circle (0.9cm);
            \end{projectxz}
            \begin{scope}[shift={(2.3,1.5,1.3)},every label/.style={fill=white,inner sep=1pt}]
                \altsphere{1.75cm}{30}
                \node[point,label={above:$c_2+e_2$}] (c2e2) at (0,0) {};
            \end{scope}
            \begin{projectxz}
                \node[point,label={below:$c_2$}] at (2.3,1.3) {};
            \end{projectxz}
            \node[color=spheremapcolor,anchor=east] at (0.3,2,2.5) {$S_2\circ\rho\circ T_2$};
        \end{scope}
        
        \begin{scope}[shift={(15.0,0)}]
            \draw[ambient] (0,-1) rectangle (1.8,3);
            \node[color=ambientcolor!80!black!80] at (0.45,-0.7) {$\R^m$};
            \node[point,label={below:$f(c_2)$}] (fc2) at (0.8,1.5) {};
        \end{scope}
        \draw[spheremapcolor,->,thick,shorten <=1mm,shorten >=1mm] (c2e2) to [bend left=10] node [pos=0.6,above] {$\Phi$} (fc2);
    \end{scope}

\end{tikzpicture}
	\caption{
		Two layers of the radial neural network used in the proof of \autoref{thm:UA-asymp-lin}.
		(Left) The compact set $K$ is covered with open balls.
		(Middle) Points close to $c_2$ (green ball) are mapped to $c_2 + e_2$, all other points are kept the same.
		(Right) In the final layer, $c_2+e_2$ is mapped to $f(c_2)$.
	}
	\label{fig:UA-asymp-lin}
\end{figure}
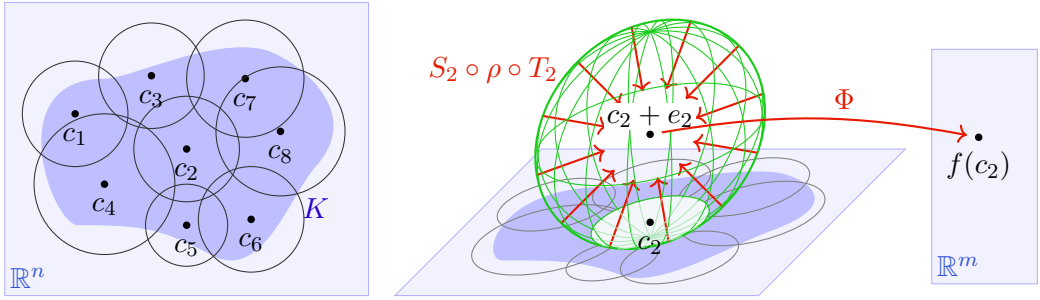

\begin{restatable}[Universal approximation]{theorem}{thmUAasymplin}
	\label{thm:UA-asymp-lin}
	Let $f : \R^n \to \R^m$ be an asymptotically affine function. For any $\epsilon >0$, there exists a compact set $K \subset \R^n$ and a function $F : \R^n \to \R^m$ such that:\vspace{-6pt}
	\begin{enumerate}
		\item $F$ is the feedforward function of a radial neural network with $N = N(f,K, \epsilon)$ layers whose hidden widths are $(n+1, n+2, \dots, n+ N)$.
		\item For any $x \in \R^n$, we have $|F(x) - f(x)| < \epsilon$. 
	\end{enumerate}
\end{restatable}
We note that the approximation in Theorem~\ref{thm:UA-asymp-lin} is valid on all of $\R^n$, not only on $K$.  To give an idea of the proof, first fix $c_1, \dots, c_N \in K$ and $r_1, \dots, r_N \in (0,1)$ as above. Let   $e_1, \dots, e_N$ be orthonormal basis vectors extending $\R^n$ to $\R^{n+N}$.   
For $i = 1, \dots, N$  define  affine maps    $T_i : \R^{n+i-1} \to \R^{n + i}$ and $S_i : \R^{n+i} \to \R^{n + i}$ by 
\begin{align*}
	T_i(z) =  z - c_i + h_i e_i \qquad & S_i(z) =  z - (1 + h_i\inv)\langle e_i, z \rangle e_i + c_i +  e_i
\end{align*}
where  $h_i = \sqrt{1 - r_i^2}$  and  $\langle e_i, z \rangle$ is the coefficient of $e_i$ in $z$.
Setting $\rho_i$ to be  Step-ReLU  (Equation \ref{eqn:step-relu}) on $\R^{n +i}$, these maps are chosen so that the composition $S_i \circ \rho_i \circ T_i$ maps the points in $B_{r_i}(c_i)$ to $c_i + e_i$, while keeping points outside this ball the same.
These maps are chosen such that $|T_i(x)| < 1$ if and only if $x \in B_{r_i}(c_i)$, $S_i(0) = c_i+e_i$, and $S_i \circ T_i = \inc_{n+i}$ is the standard inclusion $\R^{n+i} \hookrightarrow \R^{n+i+1}$.
We now describe a radial neural network with widths $(n, n+1, \dots, n+N, m)$ whose feedforward function approximates $f$.  For $i = 1, \dots, N$ the affine map from layer $i-1$ to layer $i$ is given by  $z \mapsto T_i \circ S_{i-1} (z)$, with $S_0 = \id_{\R^n}$. The activation  at each hidden layer is  {Step-ReLU}.  Let  $L$ be the affine map such that $|L-f| <\epsilon$ on $\R^n \setminus K$. The  affine map from layer $N$ to the output layer  is
\newcommand{\finalmap}{\Phi}
$\finalmap \circ S_N$
where $\finalmap : \R^{n+N} \to \R^m$ is the unique affine map determined by  $x \mapsto L(x)$ if $x \in \R^n$, and $e_i \mapsto f(c_i)- L(c_i)$. 
See Figure \ref{fig:UA-asymp-lin} for an illustration of this construction. Theorem \ref{thm:UA-asymp-lin} has the following straightforward corollary:

{\begin{corollary} 
		Radial neural networks are dense in the space of all continuous functions with respect to the topology of compact convergence, and hence  satisfy $cc$-universality.
\end{corollary}}

\subsection{Bounded width approximation}\label{subsec:bdd-width-approx}

We now turn our attention to universal approximation results using networks of bounded width. The following result is a strengthening of Theorem \ref{thm:UA-asymp-lin}.

\begin{restatable}[Bounded Width Universal Approximation]{theorem}{thmUAnplusmplusone}
	\label{thm:UA-n+m+1}
	Let $f : \R^n \to \R^m$ be an asymptotically affine function. For any $\epsilon >0$, there exists a compact set $K \subset \R^n$ and a function $F : \R^n \to \R^m$ such that:\vspace{-6pt}
	\begin{enumerate}
		\item $F$ is the feedforward function of a radial neural network with $N = N(f,K, \epsilon)$ hidden layers whose widths are all $n+m+1$.\vspace{-6pt}
		\item For any $x \in \R^n$, we have $|F(x) - f(x) | < \epsilon$. 
	\end{enumerate}
\end{restatable}
The proof, which is more involved than that of Theorem \ref{thm:UA-asymp-lin}, relies on using orthogonal dimensions to represent the domain and the range of $f$, together with an indicator dimension to distinguish the two.
We regard points in $\R^{n+m+1}$ as triples $(x,y,\theta)$ where $x \in \R^n$, $y \in \R^m$ and $\theta \in \R$.
The proof of Theorem \ref{thm:UA-n+m+1} parallels that of Theorem~\ref{thm:UA-asymp-lin},
but instead of mapping points in $B_{r_i}(c_i)$ to $c_i + e_i$, we map the points in $B_{r_i}((c_i,0,0))$ to $(0,\frac{f(c_i)- L(0)}{s},1)$, where $s$ is chosen such that different balls do not interfere.
The final layer then uses an affine map $(x,y,\theta) \mapsto L(x) + sy$,
which takes $(x,0,0)$ to $L(x)$, and $(0,\frac{f(c_i)-L(0)}{s},1)$ to $f(c_i)$.

We remark on several additional results; see Appendix~\ref{app:UA} for full statements and proofs. 
The bound of Theorem \ref{thm:UA-n+m+1} can be {strengthened} to $\max(n,m) + 1$ in the case of functions $f : K \to \R^m$ defined on a compact domain $K \subset \R^n$ (i.e., ignoring asymptotic behavior).
Furthermore, with more layers, it is possible to reduce that bound to $\max(n,m)$.


\section{Model compression}\label{sec:mod-comp}

In this section, we prove a model compression result. Specifically, we provide an algorithm which, given any radial neural network, computes a different radial neural network with smaller widths. The resulting compressed network has the same feedforward function as the original network, and hence the same value of the loss function on any batch of training data. In other words, our model compression procedure is {\it  lossless}. 
Although our algorithm is practical and explicit, it reflects more conceptual phenomena, namely, a change-of-basis action on network parameter spaces.



\subsection{Parameter space symmetries}\label{subsec:paramspace}

Suppose a fully connected network has $L$ layers and widths given by the tuple $$\bfn = (n_0, n_1, n_2, \dots, n_{L-1}, n_L).$$ In other words,  the $i$-th layer has input width $n_{i-1}$ and output width $n_i$. The parameter space  is defined as the vector space of all possible choices of parameter values. Hence, it is given by the following product of vector spaces:
\[
\Par(\bfn) = \left( \R^{n_1 \times n_0} \times \R^{n_2 \times n_1} \times \cdots \times \R^{n_L \times n_{L-1}} \right)  \times \left(\R^{n_1}  \times \R^{n_2} \times  \cdots \times \R^{n_L} \right)
\]
An element is a pair  of tuples $(\bfW, \bfb)$ where $\bfW = (W_i \in \R^{n_i \times n_{i-1}})_{i=1}^L$ are the weights and $\bfb = (b_i \in \R^{n_i})_{i=1}^L$ are the biases. 
To describe certain symmetries of the parameter space, consider the following product of orthogonal groups, with sizes  corresponding to hidden layer widths:
\[
O(\bfn^\text{hid}) = O(n_1) \times O(n_2) \times \cdots \times O(n_{L-1})
\]
There is a change-of-basis action of {this group} on the parameter space. Explicitly, the tuple of orthogonal matrices $\mathbf{Q} = (Q_\inx)\in O(\mathbf{n}^\text{\rm hid})$  transforms the parameter values  $(\bfW, \bfb) $
to  $ \mathbf{Q} \cdot  \bfW :=   \left( Q_\inx  W_\inx   Q_{\inx-1}\inv \right)$ and $  \mathbf{Q} \cdot  \bfb :=   \left( Q_\inx  b_i\right)$, where we set $Q_0$ and $Q_L$ to be identity matrices. 



\subsection{Model compression}\label{subsec:mod-comp}

In order to state the compression result, we first define the reduced widths. Namely,   the {reduction}  $\bfnred = (\nred_0, \nred_1,    \dots, \nred_L)$ of a widths vector $\mathbf n$ is defined recursively by setting $\nred_0 = n_0 $, then $\nred_{i} = \min( n_i  , \nred_{i-1} + 1  )$ for $i = 1, \dots, L-1$, and finally $\nred_L = n_L$.  For a tuple  $\boldrho = \left(\rho_i : \R^{n_i} \to \R^{n_i} \right)_{i=1}^L$ of radial rescaling functions, 
we write $\brhored = \left(\rhored_i : \R^{\nred_i} \to \R^{\nred_i}\right)$ for the corresponding  tuple of restrictions, which are all radial rescaling functions.
The following result relies on Algorithm \ref{alg:QR-mod-comp} below. 


\begin{restatable}{theorem}{thmModelCompression}
	\label{thm:mod-comp}
	Let $(\bfW, \bfb, \boldrho  )$ be a radial neural network with widths $\bfn$. Let $\bfWred$ and $\bfbred$ be the  weights and biases of the compressed network produced by Algorithm \ref{alg:QR-mod-comp}.
	The feedforward function of the original network  $(\bfW, \bfb, \boldrho  )$ coincides with that of the compressed network   $(\bfWred, \bfbred, \brhored  )$.
\end{restatable}


\begin{algorithm}\label{alg:QR-mod-comp}
		\caption{QR Model Compression (\texttt{QR-compress})}
		\BlankLine
\hrule
\BlankLine
	\SetKwFunction{QRdecompCom}{QR-decomp}
	\SetKwFunction{QRdecompRed}{QR-decomp}
	\SetKwInOut{Input}{input}
	\SetKwInOut{Output}{output}
	\SetKwInOut{Initialize}{initialize}
	\DontPrintSemicolon
	\Input{$\bfW, \bfb \in \Par(\bfn)$}
	\Output{$\mathbf{Q} \in O(\bfn^\text{\rm hid})$ and $\bfWred, \bfbred \in \Par(\bfnred)$}
	\BlankLine
	
	$\bfQ, \bfWred, \bfbred \gets [\ ], [\ ], [\ ]$  
	\tcp*[r]{initialize output lists}
	
	$A_1 \gets \begin{bmatrix} b_1 & W_1 \end{bmatrix}$\: 
	\tcp*[r]{matrix of size $n_1 \times (n_0 + 1)$}
	
	\For(\tcp*[r]{iterate through layers \vspace{-\baselineskip}})
	{$ \inx \leftarrow 1$ \KwTo $L-1$   }{ 
		$Q_\inx, R_\inx \gets $ \QRdecompCom{$A_\inx$ \ , \  \text{\tt mode = `complete'}} \tcp*[r]{$A_\inx = Q_\inx  \Inc_i  R_\inx$}
		
		Append $Q_\inx$ to $\bfQ$\;
		
		Append first column of $R_i$ to $\bfbred$
		\tcp*[r]{reduced bias for layer $i$}
		
		Append remainder of $R_i$ to $\bfWred$ 
		\tcp*[r]{reduced weights for layer $i$}	
		
		Set $A_{\inx+1} \gets  \begin{bmatrix} b_{i+1}  &  W_{i+1}  Q_i  \Inc_i  \end{bmatrix}$ 
		\tcp*[r]{matrix of size {\footnotesize $n_{i+1} \times  (\nred_{i} + 1)$} } 	
	}
	Append the first column of $A_L$ to $\bfbred$
	\tcp*[r]{reduced bias for last layer}
	Append the remainder of $A_L$  to $\bfWred$
	\tcp*[r]{reduced weights for last layer}
	\KwRet $\mathbf{Q}$, $\bfWred$, $\bfbred$
	\BlankLine
	\hrule
	\BlankLine
\textit{Notation}: The inclusion matrix $\Inc_i \in \R^{n_i \times \nred_i}$ has ones along the main diagonal and zeros elsewhere. 
The method \texttt{QR-decomp} with \texttt{mode = `complete'} computes the complete QR decomposition of the $n_\inx \times (1 +  \nred_{\inx-1})$ matrix $A_\inx$ as  $Q_\inx \Inc_i  R_\inx$ where $Q_\inx \in O(n_\inx)$ and $R_\inx$ is upper-triangular of size $\nred_ \inx \times (1+\nred_{i-1})$. The definition of $\nred_\inx$  implies that either  $\nred_\inx = \nred_{\inx-1} + 1$  or $\nred_\inx = n_\inx$.  The matrix $R_\inx$ is of size $\nred_ \inx \times \nred_{\inx}$ in the former case and of size $n_\inx \times  (1 +  \nred_{\inx-1})$ in the latter case.

\end{algorithm}


We note that the tuple of matrices $\bfQ$ produced by Algorithm \ref{alg:QR-mod-comp} does not feature in the statement of Theorem \ref{thm:mod-comp}, but is important in the proof (which appears in Appendix \ref{app:mod-comp}). Namely,  an induction argument shows that the $i$-th partial feedforward function of the original and reduced models are related via the matrices $Q_i$ and $\Inc_i$. A crucial ingredient in the proof is that radial rescaling activations commute with orthogonal transformations.

\begin{figure}[b]
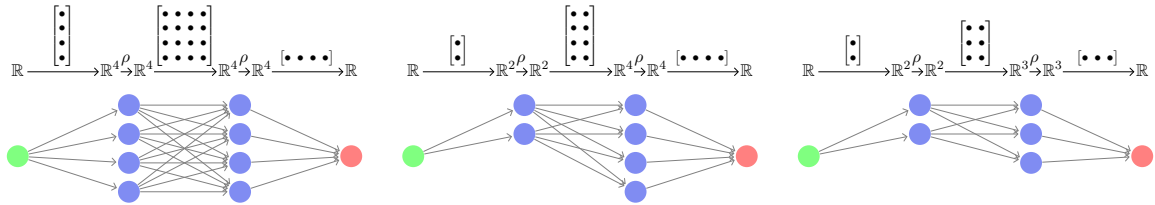

	\centering%
	\def\layersep{2.5cm}%
\def\neuronsep{0.65cm}%
\def\rholayersep{0.75cm}%
%
\newenvironment{compactbmatrix}{%
  \begingroup%
  \setlength\arraycolsep{1.9pt}%
  \begin{bmatrix}%
}{
  \end{bmatrix}%
  \endgroup%
}%
\newcommand{\NNQuiverDiagram}[5]{
  \resizebox{0.31\textwidth}{!}{%
    \begin{tikzpicture}[baseline,->,>=angle 90]

      \begin{scope}[yshift=0.5cm,anchor=base,nodes={text height=3mm,text depth=1mm,inner sep=2pt}]
        \node (A) at (0*\layersep,0)                  {$\displaystyle\mathbb{R}$};
        \node (B) at (1*\layersep-0.6*\rholayersep,0) {$\displaystyle\mathbb{R}^{#1}$\hspace*{-2pt}};
        \node (C) at (1*\layersep+0.4*\rholayersep,0) {$\displaystyle\mathbb{R}^{#1}$\hspace*{-2pt}};
        \node (D) at (2*\layersep-0.4*\rholayersep,0) {$\displaystyle\mathbb{R}^{#2}$\hspace*{-2pt}};
        \node (E) at (2*\layersep+0.6*\rholayersep,0) {$\displaystyle\mathbb{R}^{#2}$\hspace*{-2pt}};
        \node (F) at (3*\layersep,0)                  {$\displaystyle\mathbb{R}$};
      \end{scope}

      \path[->,>=angle 90,shorten >=-1pt]
      (A) edge node[above]{
          \scriptsize$#3$
      } (B)
      (B) edge node[above]{\raisebox{0mm}{$\rho$}} (C)
      (C) edge node[above]{
          \scriptsize$#4$
      } (D)
      (D) edge node[above]{\raisebox{0mm}{$\rho$}}  (E)
      (E) edge node[above]{
          \scriptsize$#5$
      } (F)
      ;
      

      \tikzstyle{every pin edge}=[<-,shorten <=1pt]
      \tikzstyle{neuron}=[circle,fill=black!25,minimum size=14pt,inner sep=0pt]
      \tikzstyle{input neuron}=[neuron, fill=green!50];
      \tikzstyle{output neuron}=[neuron, fill=red!50];
      \tikzstyle{hidden neuron}=[neuron, fill=green!10!blue!50];
      \tikzstyle{annot} = [text width=4em, text centered]

      \foreach \name / \y in {1,...,1}
          \node[input neuron] (I-\name) at (0,-2*\neuronsep) {};

      \foreach \name / \y in {1,...,#1}
          \path[yshift=0.5cm]
              node[hidden neuron] (H-\name) at (\layersep,{-(\y+0*(4-#1)/2)*\neuronsep}) {};

      \foreach \name / \y in {1,...,#2}
          \path[yshift=0.5cm]
              node[hidden neuron] (H2-\name) at (2*\layersep,{-(\y+0*(4-#2)/2)*\neuronsep}) {};

      \node[output neuron] (O) at (3*\layersep, -2*\neuronsep) {};

      \begin{scope}[shorten >=1pt,draw=black!50]
          \foreach \source in {1,...,1}
              \foreach \dest in {1,...,#1}
                  \path (I-\source) edge (H-\dest);
                  
          \foreach \source in {1,...,#1}
              \foreach \dest in {1,...,#2}
                  \path (H-\source) edge (H2-\dest);

          \foreach \source in {1,...,#2}
              \path (H2-\source) edge (O);
      \end{scope}
      
    \end{tikzpicture}%
  }%
}%
\NNQuiverDiagram{4}{4}{
    \begin{compactbmatrix}
        \bullet \\
        \bullet \\
        \bullet \\
        \bullet
    \end{compactbmatrix}
}{
    \begin{compactbmatrix}
        \bullet & \bullet & \bullet & \bullet \\
        \bullet & \bullet & \bullet & \bullet \\
        \bullet & \bullet & \bullet & \bullet \\
        \bullet & \bullet & \bullet & \bullet
    \end{compactbmatrix}
}{
    \begin{compactbmatrix}
        \bullet & \bullet & \bullet & \bullet
    \end{compactbmatrix}
}%
%
	\hfill%
	\input{figs/NNquiver2}%
	\hfill%
	\input{figs/NNquiver3}%
	\caption{Model compression in 3 steps.  Layer widths can be iteratively reduced to 1 greater than the previous. The number of trainable parameters reduces from 33 to 17.}
	\label{fig:dim_red}
\end{figure}

\begin{example}
	Suppose the  widths of a radial neural network are given by $ (1, 8, 16, 8, 1 )$. Then it has $\sum_{i=1}^4 (n_{i-1} +1)n_i = 305$ trainable parameters. The reduced network has widths $(1, 2, 3, 4, 1)$ and $\sum_{i=1}^4 (\nred_{i-1} + 1)(\nred_i) = 34$ trainable parameters. Another example appears in Figure \ref{fig:dim_red}.
\end{example}

\section{Projected gradient descent}\label{subsec:proj-gd}

The typical use case for model compression algorithms is to produce a smaller version of the fully trained model which can be deployed to make inference more efficient. It is also  worth considering whether compression can be used to accelerate training.  For example, for some compression algorithms, the compressed and full models have the same feedforward function after a step of gradient descent is applied to each, and so one can  compress before training and still reach the same minimum. 
Unfortunately, in the context of radial neural networks, compression using Algorithm \ref{alg:QR-mod-comp} and then training does not necessarily give the same result as training and then compression (see Appendix \ref{appsubsec:ex-131} for a counterexample).  However, \texttt{QR-compress} does lead to a precise mathematical relationship between optimization of the two models:
the loss of the compressed model after one step of gradient descent  is equivalent to the  loss of (a transformed version of) the original model after one  step of {projected gradient descent}. Proofs appear in Appendix \ref{app:proj-gd}.


To state our results, fix   widths $\mathbf n$ and   radial rescaling functions $\boldrho$ as above. The loss function $\mathcal L :  \Par(\bfn)\rightarrow \R$ associated to  a batch of training data  
is defined as taking parameter values $ (\mathbf{W} , \bfb ) $ to the sum 
$\sum_j \mathcal C(F ( x_j),  y_j)$
where $\mathcal C$
is a cost function on the output space, $F = F_{(\bfW, \bfb, \boldrho)}$ is the feedforward of the radial neural network with the specified parameters, and $(x_j, y_j) \in \R^{n_0} \times \R^{n_L}$ are the data points. Similarly, we have a loss function  $\mathcal L_\text{\rm red}$ on the parameter space $\Par(\bfnred)$ with reduced widths vector. For any learning rate $\eta >0$,  we obtain   gradient descent maps:
\begin{align*}
	\gamma : \Par(\bfn) & \to \Par(\bfn);  \qquad &(\bfW, \bfb) &\mapsto (\bfW, \bfb)  -  \eta \nabla_{(\bfW, \bfb) } \mathcal L \\
	\gamma_{\text{\rm red}} : \Par(\bfnred) &\to \Par(\bfnred);    & (\bfV, \bfc )&\mapsto (\bfV, \bfc ) -  \eta \nabla_{(\bfV, \bfc )} \mathcal L_\text{\rm red} \\
	\gamma_{\text{\rm proj}} : \Par(\bfn) &\to \Par(\bfn) ; \qquad  & (\bfW, \bfb)  &\mapsto  \Proj\left( \gamma(\bfW, \bfb) \right)
\end{align*}
where the last is the  {\it  projected gradient descent} map on $\Par(\bfn)$. The map $\Proj$ zeroes out  all entries in the bottom left $(n_{i} - \nred_{i}) \times \nred_{i-1}$ submatrix of $W_i - \nabla_{W_i}\mathcal L$, and  the bottom $(n_i - \nred_i)$ entries in $b_i - \nabla_{b_i} \mathcal L$, for each $i$. Schematically:
\[
\small   {W_i} - \nabla_{W_i} \mathcal L= \begin{bmatrix}
	* & * \\
	* & *
\end{bmatrix}   \mapsto    \begin{bmatrix}
	* & * \\
	0 & *
\end{bmatrix} , 
\qquad 
{b_i} - \nabla_{b_i} \mathcal L= \begin{bmatrix}
	*  \\
	*
\end{bmatrix}   \mapsto    \begin{bmatrix}
	*  \\
	0 
\end{bmatrix} 
\]
{ To state the following theorem,  recall that, applying Algorithm \ref{alg:QR-mod-comp} to parameters $(\bfW, \bfb)$, we obtain the reduced model $(\bfWred, \bfbred )$ and an orthogonal parameter symmetry $\bfQ$. We consider, for $k \geq 0$, the $k$-fold composition $\gamma^k = \gamma \circ \gamma \circ \cdots \circ \gamma$ and similarly for  $\gamma_{\text{\rm red}}$ and $\gamma_{\text{\rm red}}$.} 


\begin{theorem}\label{thm:proj-gd}
	Let $\bfWred, \bfbred, \bfQ = \text{\tt{QR-compress}}(\bfW, \bfb)$ be the outputs of Algorithm \ref{alg:QR-mod-comp} applied to $(\bfW, \bfb) \in \Par(\bfn)$. Set  $\bfU = \bfQ\inv \cdot \left( \bfW , \bfb \right) -  ( \bfWred,  \bfbred)$. For any $k \geq 0$, we  have: 
		\[
		\gamma^k( \bfW, \bfb) = \bfQ \cdot \gamma^k( \bfQ\inv \cdot \left( \bfW , \bfb \right)) \qquad \qquad \gamma_\text{\rm proj}^k (\bfQ\inv \cdot \left( \bfW , \bfb \right)) =   \gamma_{\text{\rm red}}^k (\bfWred, \bfbred)  + \bfU.
		\]
\end{theorem} 


We conclude that gradient descent with initial values $(\bfW, \bfb)$ is equivalent to gradient descent with initial values $ \bfQ\inv \cdot \left( \bfW , \bfb \right)$ since at any stage we can apply $\bfQ^{\pm 1}$ to move from one to the other (using the  action from Section \ref{subsec:paramspace}). Furthermore, projected gradient descent with initial values $ \bfQ\inv \cdot \left( \bfW , \bfb \right)$ is equivalent to gradient descent on $\Par(\bfnred)$ with initial values $(\bfWred, \bfbred)$ since at any stage we can move from one to the other by $\pm\bfU$. {Neither $\bfQ$ nor $\bfU$ depends on $k$.}
\section{Experiments}\label{sec:exp}

In addition to our theoretical results, we provide an implementation of  
Algorithm \ref{alg:QR-mod-comp} in order to validate the claims of Theorems \ref{thm:mod-comp} and \ref{thm:proj-gd} empirically, as well as a demonstration that a radial network outperforms a MLP on a noisy image recovery task.
Full experimental details are in Appendix \ref{app:exp}.

\begin{figure}[t]
	\centering
	\includegraphics[width=0.4\textwidth]{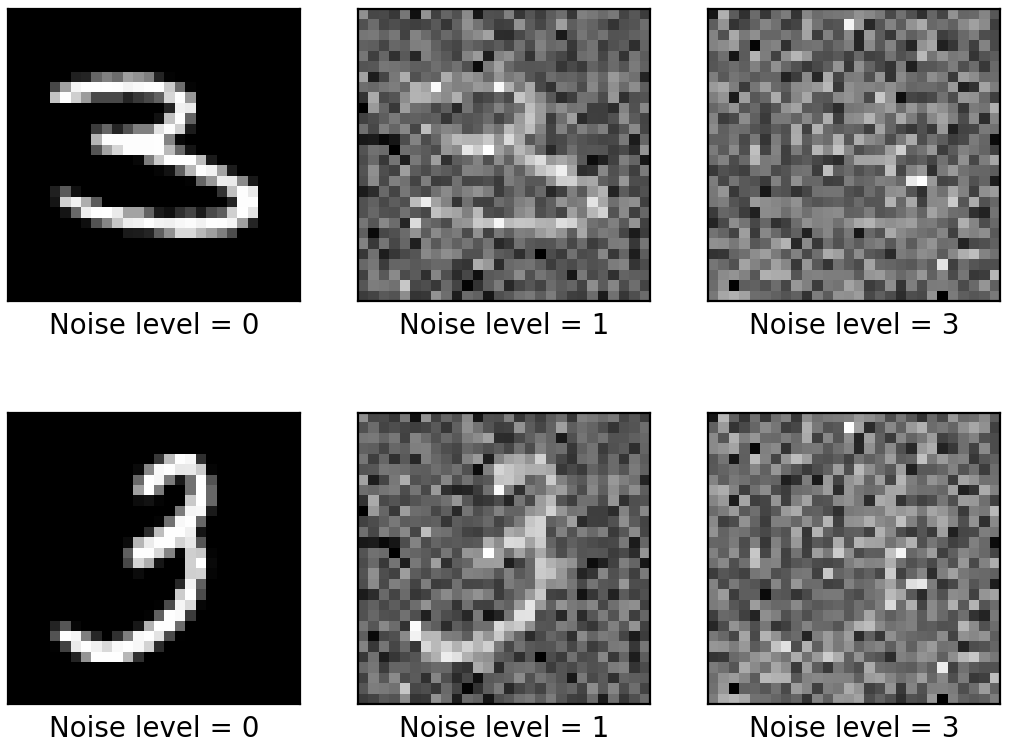}
	\qquad  \qquad
	\includegraphics[width=0.35\textwidth]{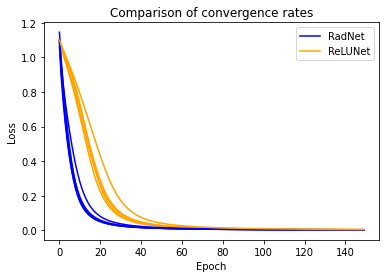}
	\caption{(Left) Different levels of noise. (Right) Training five Step-ReLU radial networks and five ReLU MLPs on data with \texttt{n}=3 original images, \texttt{m}=100 noisy copies of each.
	}
	\label{fig:noisy-threes}
\end{figure}

\begin{enumerate}
    \item {\bf Empirical verification of Theorem \ref{thm:mod-comp}.}\label{subsec:exp-QR-thm}
We learn the function $f(x) = e^{-x^2}$ from samples using a radial neural network
with widths $\bfn = (1,6,7,1)$ and activation the radial shifted sigmoid $h(x) = 1/(1+e^{-x + s})$.  Applying \texttt{QR-compress} gives a compressed radial neural network with widths $\bfn^{\mathrm{red}} = (1,2,3,1)$.  Theorem \ref{thm:mod-comp} implies that the respective neural functions $F$ and $F_\text{\rm red}$ are equal.  Over 10 random initializations, the mean absolute error is negligible up to machine precision: $(1/N) \sum_{j} |F(x_j) - F_\text{\rm red}(x_j)| = 1.31 \cdot 10^{-8} \pm 4.45 \cdot 10^{-9}$.

    \item {\bf Empirical verification of Theorem \ref{thm:proj-gd}.}\label{subsec:exp-projGD-thm}
The claim  is that training the transformed model with  parameters $\bfQ\inv \cdot (\bfW, \bfb)$
and objective $\mathcal{L}$ by projected gradient descent coincides with training the reduced model with  parameters $(\bfWred, \bfbred)$ 
and objective $\mathcal{L}_{\mathrm{red}}$ by usual gradient descent. We verified this on synthetic data as above. Over 10 random initializations, the loss functions after training match: $|\mathcal{L}-\mathcal{L}_{\mathrm{red}}| = 4.02 \cdot 10^{-9} \pm 7.01 \cdot 10^{-9}$.



\item {\bf The compressed model trains faster.}  \label{subsec:exp-faster}
Our compression method may be applied before training to produce a smaller model class which \emph{trains} faster without sacrificing accuracy. We demonstrate this in learning the function $\R^2 \to \R^2$ sending $(t_1, t_2)$ to $(e^{-t_1^2}, e^{-t_2^2})$ using a radial neural network with widths $(2,16,64, 128, 16, 2)$ and activation the radial sigmoid $h(r) = 1/(1+e^{-r})$.  Applying  \texttt{QR-compress}  gives a compressed network  with widths $\bfn^{\mathrm{red}} = (2,3,4,5,6,2)$. We trained both models until the training loss was $\leq0.01$.  Over 10 random initializations on our system, the reduced network trained in $15.32 \pm 2.53$ seconds and the original network trained in $31.24 \pm 4.55$ seconds.

\item {\bf Noisy image recovery.} A Step-ReLU radial network performs better than an otherwise comparable network with pointwise ReLU on a noisy image recovery task. Using samples of MNIST with significant added noise, the network must identify from which original sample the noisy sample derives (see Figure \ref{fig:noisy-threes}).  
We observe that the radial network 1) is able to obtain a better fit, 2) has faster convergence, and 3) generalizes better than the pointwise ReLU.  We hypothesize the radial nature of the random noise makes radial networks well-adapted to the task. Our data takes \texttt{n} = 3 original MNIST images with the same label, and produces \texttt{m} = 100 noisy images for each, with a 240 train / 60 test split. Over 10 trials, each training for 150 epochs, the radial network achieves training loss 0.00256 $\pm 3.074  \cdot 10^{-1}$ with accuracy 1 $\pm$ 0, while the ReLU MLP has training loss 0.00393 $\pm 3.992 \cdot 10^{-4}$ with accuracy 1 $\pm$ 0. On the test set, the radial network has loss 0.00266 $\pm 3.749  \cdot 10^{-4}$ with accuracy 1 $\pm$ 0, while the  ReLU MLP has loss 0.00413 $\pm 4.442  \cdot 10^{-4}$ with accuracy 1 $\pm$ 0.
The convergence rates are illustrated in Figure \ref{fig:noisy-threes}, with the radial network outperforming the ReLU MLP, and 150 epochs are sufficient for all methods to converge.


\end{enumerate}

\section{Conclusions and Discussion}\label{sec:conclusions}

This paper demonstrates that radial neural networks are universal approximators and that their parameter spaces exhibit a rich symmetry group, leading to a model compression algorithm. The results of this work combine to build a theoretical foundation for the use of radial neural networks, and suggest that radial neural networks hold promise for wider practical applicability. Furthermore, this work makes an argument for considering non-pointwise nonlinearities in neural networks.

There are two main limitations of our results, each providing an opportunity for future work. First, our universal approximation constructions currently work only for Step-ReLU radial rescaling radial activations; it would be desirable to generalize to other activations. Additionally, Theorem \ref{thm:mod-comp} achieves compression only for networks whose widths satisfy  $n_i > n_{i-1}+1$ for some $i$. 
Networks which do not have increasing widths anywhere, such as encoders, would not be compressible. 

Further extensions of this work include: First, little is currently known about the stability properties of radial neural networks during training, as well as their sensitivity to initialization. Second, radial rescaling activations provide an extreme case of symmetry; there may be benefits to combining radial and pointwise activations within a single network, for example, through `block' radial rescaling functions. {Our techniques may yield weaker compression properties for more general radial basis functions networks;  radial neural networks may be the most compressible such networks.}
Third, the parameter space symmetries may provide a key ingredient in analyzing the gradient flow dynamics of radial neural networks and computation of conserved quantities. 
Fourth, radial rescaling activations can be used within convolutional or group-equivariant NNs. 
%
%
Finally, based on the theoretical advantages and experiments laid out in this paper, future empirical work will  further explore  applications in which we expect radial networks to outperform alternate methods. Such potential applications include data spaces with circular or distance-based class boundaries.

\acks{We would like to thank Avraham Aizenbud, Marco Antonio Armenta, Alex Kolmus, Niklas
	Smedemark-Margulies, Jan-Willem van de Meent, and Rose Yu for insightful discussions,
	comments, and questions. This work was (partially) funded by the NWO under the
	CORTEX project (NWA.1160.18.316) and NSF grant \#2134178. Robin Walters is supported
	by the Roux Institute and the Harold Alfond Foundation.}

\clearpage

\bibliography{ref}

\clearpage

\appendix

\section{Organization of the appendices}\label{app:sum}

This paper is a contribution to the mathematical foundations of machine learning, and our results are motivated by expanding the applicability and performance of neural networks. At the same time, we give precise mathematical formulations of our results and proofs. The purposes of these appendices are several:
\begin{enumerate}
	\item To clarify the mathematical conventions and terminology, thus making the paper more accessible.
	\item To provide full proofs of the main results.
	\item To develop context around various construction appearing in the main text.
	\item  To discuss in detail examples, special cases, and generalizations of our results.
	\item To specify implementation details for the experiments.
\end{enumerate}

We now give a summary of the contents of the appendices.  

Appendix \ref{app:UA} contains proofs the universal approximation results (Theorems \ref{thm:UA-asymp-lin} and \ref{thm:UA-n+m+1}) stated in Section \ref{sec:UA} of the main text, as well as proofs of additional bounded width results. The proofs use notation given in Appendix \ref{app:UA-notation}, and rely on  preliminary topological considerations given in Appendix \ref{subsec:topology}. 

In Appendix \ref{app:mod-comp}, we give a proof of the model compression result given in Theorem \ref{thm:mod-comp}, which appears in Section \ref{sec:mod-comp}. For clarity and background we begin the appendix with a discussion of the version of the QR decomposition  relevant for our purposes (Appendix \ref{subsec:qr}). We also establish elementary properties of radial rescaling activations (Appendix \ref{subsec:radial-properties}). 

The focus of  Appendix \ref{app:proj-gd} is projected gradient descent, elaborating on Section \ref{subsec:proj-gd}. We first prove a result on the interaction of gradient descent and orthogonal transformations (Appendix \ref{app:gd-orth}), before formulating projected gradient descent in more detail (Appendix \ref{app:set-up}), and introducing the so-called interpolating space (Appendix \ref{appsubsec:interpolating}). 
We restate Theorem \ref{thm:proj-gd} in more convenient notation (Appendix \ref{appsubsec:projGD-QR}) before proceeding to the proof (Appendix \ref{appsubsec:proof-projGD}).

Appendix \ref{app:exp} contains implementation details for the experiments summarized in Section \ref{sec:exp}. Several of our implementations use shifted radial rescaling activations, which we formulate in Appendix \ref{appsubsec:withshifts}. 

{Appendix \ref{app:rbfns} explains the connection between our constructions and radial basis functions networks. While radial neural networks turn out to be a specific type of radial basis functions network, our universality results are not implied by those for general radial basis functions networks. 
}

\section{Universal approximation proofs and additional results}\label{app:UA}

In this section, we provide full proofs of the universal approximation (UA) results for radial neural networks, as stated in Section \ref{sec:UA}. In order to do so, we first clarify our notational conventions (Appendix \ref{app:UA-notation}), and collect basic topological results (Appendix \ref{subsec:topology}). 

\subsection{Notation}\label{app:UA-notation}

Recall that, for a point $c$ in the Euclidean space $\R^n$ and a positive real number $r$, we denote the $r$-ball around $c$ by $B_r(c) = \{ x \in \R^n  \ | \ |x-c| < r\}$. All networks in this section have the Step-ReLU radial rescaling activation function, defined as:
\begin{align*}
	\rho : \R^{n} & \longrightarrow \R^{n}, \qquad	z \longmapsto 
	\begin{cases} 	z & \qquad \text{\rm if $|z| \geq 1$} \\
		0 & \qquad \text{\rm otherwise} 	
	\end{cases} 	
\end{align*}
Throughout, $\circ$ denotes the composition of functions. We identify a linear map with a corresponding matrix (in the standard bases). In the case of linear maps, the operation $\circ$ can be be identified with matrix multiplication.  Recall also that an affine map $L : \R^n \to \R^m$ is one of the from $L(x)= Ax + b$ for a matrix $A \in \R^{m \times n}$ and $b\in \R^m$.

\subsection{Topology}\label{subsec:topology}

Let $K$ be a compact subset of $\R^n$ and let $ f: K \to \R^m$ be a continuous function.

\begin{lemma}\label{lem:def-N}
	For any $\epsilon > 0$, there exist $c_1, \dots, c_N \in K$ and $r_1, \dots, r_N \in (0,1)$ such that, first, the union of the balls $B_{r_i} (c_i)$ covers $K$; second, for all $i$, we have $f \left(B_{r_i} (c_i) \cap K \right) \subseteq  B_{\epsilon}(f(c_i))$.
\end{lemma}

\begin{proof}
	The continuity of $f$ implies that for each $c \in K$, there  exists $r = r_c$ such that $f(B_{r_c}(c) \cap K) \subseteq B_\epsilon(f(c))$. The subsets $B_{r_c}(c) \cap K$ form an open cover of $K$. The compactness of $K$ implies that there is a finite subcover. The result follows. 
\end{proof}


We also prove a variation of Lemma \ref{lem:def-N} that additionally guarantees that none of the balls in the cover of $K$ contains the center point of another ball.

\begin{lemma}\label{lem:def-M}
	For any $\epsilon > 0$, there exist $c_1, \dots, c_M \in K$ and $r_1, \dots, r_M \in (0,1)$ such that, first, the union of the balls $B_{r_i} (c_i)$ covers $K$; second, for all $i$, we have $f \left(B_{r_i} (c_i) \right) \subseteq  B_{\epsilon}(f(c_i))$; and, third, $|c_i - c_j| \geq r_i$. 
\end{lemma}

\begin{proof}
	Because $f$ is continuous on a compact domain, it is uniformly continuous.
	So, there exists $r > 0$ such that $f(B_{r}(c) \cap K) \subseteq B_\epsilon(f(c))$ for each $c \in K$.
	Because $K$ is compact it has a finite volume, and so does $B_{r/2}(K) = \bigcup_{c \in K} B_{r/2}(c)$.
	Hence, there exists a finite maximal packing of $B_{r/2}(K)$ with balls of radius $r/2$.
	That is, a collection $c_1,\dotsc,c_M \in B_{r/2}(K)$ such that, for all $i$, $B_{r/2}(c_i) \subseteq B_{r/2}(K)$ and, for all $j \neq i$, $B_{r/2}(c_i) \cap B_{r/2}(c_j) = \emptyset$.
	The first condition implies that $c_i \in K$.
	The second condition implies that $|c_i - c_j| \geq r$.
	Finally, we argue that $K \subseteq \bigcup_{i=1}^M B_{r}(c_i)$. To see this, suppose, for a contradiction, that $x \in K$ does not belong to $\bigcup_{i=1}^M B_{r}(c_i)$. Then $B_{r/2}(c_i) \cap B_{r/2}(x) = \emptyset$, and $x$ could be added to the packing, which contradicts the fact that the packing was chosen to be maximal.
	So the union of the balls $B_{r}(c_i)$ covers $K$.
\end{proof}



We turn our attention to the minimal choices of $N$ and $M$ in Lemmas \ref{lem:def-N} and \ref{lem:def-M}. 

\begin{definition}\label{def:NM}
	Given $f: K \to \R^m$ continuous and $\epsilon >0$, let $N(f,K, \epsilon)$ be the minimal choice of $N$ in Lemma \ref{lem:def-N}, and let $M(f, K, \epsilon)$ be the minimal choice of $M$ in Lemma \ref{lem:def-M}. 
\end{definition}

Observe that $M(f,K, \epsilon) \geq N(f, K, \epsilon)$.  In many cases, it is possible to give explicit bounds for the constants $N(f,K, \epsilon)$ and $M(f,K, \epsilon)$. As an illustration, we give the argument in the case that $K$ is the closed unit cube in $\R^n$ and $f : K \to \R^m$ is Lipschtiz continuous. 

\begin{proposition}
	\label{thm:bound-N}
	Let $K = [0,1]^n \subset \R^n$ be the (closed) unit cube and let $f : K \to \R^m$ be Lipschitz continuous with Lipschitz constant $R$. 
	For any $\epsilon >0$,  we have:
	\[
	N(f,K, \epsilon ) \leq \left\lceil \frac{R \sqrt{n}}{2\epsilon} \right\rceil^n
	\qquad \text{and} \qquad 
	M(f,K,\epsilon) \leq
	\frac{\Gamma(n/2+1)}{\pi^{n/2}} \left(2 + \frac{2R}{\epsilon}\right)^n.
	\]
\end{proposition}

\begin{proof}
	For the first inequality, observe that the unit cube can be covered with $\Bigl\lceil \frac{R \sqrt{n}}{2\epsilon} \Bigr\rceil^n$ cubes of side length $\frac{2\epsilon}{R\sqrt{n}}$. Each cube is contained in a ball of radius $\frac{\epsilon}{R}$ centered at the center of the cube.  (In general, a cube of side length $a$ in $\R^n$ is contained in a ball of radius $\frac{a \sqrt{n}}{2}$.)
	Lipschitz continuity implies that, for all $x,x' \in K$, if $|x-x'| < \epsilon/R$ then $|f(x) - f(x')| \leq R |x- x'| < \epsilon$.
	
	\newcommand{\vol}{\text{vol}}
	
	For the second inequality, let $r=\epsilon/R$.
	Lipschitz continuity implies that, for all $x,x' \in K$, if $|x-x'| < r$ then $|f(x) - f(x')| \leq R |x-x'| < \epsilon$.
	The $n$-dimensional volume of the set of points with distance at most $r/2$ to the unit cube is $\vol(B_{r/2}(K)) \le (1+r)^n$.
	The volume of a ball with radius $r/2$ is $\vol(B_{r/2}(0)) = \frac{\pi^{n/2}}{\Gamma(n/2+1)}(r/2)^n$.
	Hence, any packing of $B_{r/2}(K)$ with balls of radius $r/2$ consists of at most \[\frac{\vol(B_{r/2}(K))}{\vol(B_{r/2}(0))} \le \frac{\Gamma(n/2+1)}{\pi^{n/2}} \left(2 + \frac{2R}{\epsilon}\right)^n
	\]
	such balls.
	So there also exists a maximal packing with at most that many balls. This packing can be used in the proof of \autoref{lem:def-M}, which implies that it is a bound on $M(f,K,\epsilon)$.
\end{proof}

We note in passing that any differentiable function $f : K \to \R^n$ on a compact subset $K$ of $\R^n$ is Lipschitz continuous. Indeed, the compactness of $K$ implies that there exists $R$ such that $|f^\prime (x)| \leq R$ for all $x \in K$. Then one can take $R$ to be the Lipschitz constant of $f$. 

\subsection{Proof of Theorem \ref{thm:UA-asymp-lin}: UA for asymptotically affine functions}\label{app:asymp-lin}

In this section, we restate and prove Theorem \ref{thm:UA-asymp-lin}, which proves that radial neural networks are universal approximators of asymptotically affine functions. We recall the definition of such functions:

\begin{definition}\label{def:asymp-lin-alt}
	A function $f : \R^n \to \R^m$ is \emph{asymptotically affine} if there exists an affine function $L : \R^n \to \R^m$ such that, for all $\epsilon > 0$, there exists  a compact set $K \subset \R^n$ such that $|L(x) - f(x)| < \epsilon$ for all $x \in \R^n \setminus K$. We say that $L$ is the limit of $f$. 
\end{definition}

\begin{remark} An {\it asymptotically linear} function is defined in the same way, 
	except $L$ is taken to be linear (i.e., given just by applying matrix multiplication without translation). Hence any asymptotically linear function is in particular an asymptotically affine function, and Theorem \ref{thm:UA-asymp-lin} applies to asymptotically linear functions as well.
\end{remark}

Given an asymptotically affine function $f : \R^n \to \R^m$ and $\epsilon >0$, let $K$ be a compact set as in Definition \ref{def:asymp-lin-alt}. We apply Lemma~\ref{lem:def-N} to the restriction $f \vert_K$ of $f$ to $K$ and produce a minimal constant $N = N(f\vert_K,K,\epsilon)$ as in Definition \ref{def:NM}. We write simply $N(f,K,\epsilon)$ for this constant.

\thmUAasymplin*
\widthsublemmasref{thm:UA-asymp-lin}

\begin{proof}
	By the hypothesis on $f$, there exists an affine function $L : \R^n \to \R^m$ and a compact set $K \subset \R^n$ such that $|L(x) - f(x)| < \epsilon$ for all $x \in \R^n \setminus K$.
	Abbreviate $N(f,K,\epsilon)$ by $N$. As in Lemma~\ref{lem:def-N}, fix $c_1, \dots, c_N \in K$ and $r_1, \dots, r_N \in (0,1)$ such that, first, the union of the balls $B_{r_i} (c_i)$ covers $K$ and, second, for all $i$, we have $f \left(B_{r_i} (c_i) \right) \subseteq  B_{\epsilon}(f(c_i))$. 
	Let $U =  \bigcup_{i=1}^N B_{r_i}(c_i)$, so that $K \subset U$.  Define $F : \R^n \to \R^m$ as:
	\[
	F(x) = \begin{cases}
		L(x) &\qquad \text{\rm if $x \notin U$} \\
		f(c_j) & \qquad \text{\rm where $j$ is the smallest index with $x \in B_{r_j} (c_j)$}
	\end{cases}
	\]
	If  $x \notin U$, then $|F(x) - f(x)| = |L(x) - f(x)| < \epsilon$. Hence suppose  $x \in U$. Let $j$ be the smallest index such that $x\in B_{r_j}(c_j)$. Then $F(x) = f(c_j)$, and, by the choice of $r_j$, we have:
	\[|F(x) - f(x)| = |f(c_j) - f(x)| < \epsilon.\]
	
	We proceed to show that $F$ is the feedforward function of a radial neural network. 	
	Let  $e_1, \dots, e_N$ be orthonormal basis vectors extending $\R^n$ to $\R^{n+N}$.   We regard each $\R^{n+i-1}$ as a subspace of $\R^{n+i}$ by embedding into the first $n+i-1$ coordinates. 
	For $i = 1, \dots, N$, we set $h_i = \sqrt{1 - r_i^2}$ and  define the following affine transformations:
	\begin{align*}
		T_i : \R^{n+i-1} &\to \R^{n + i} \qquad & S_i : \R^{n+i} & \to \R^{n + i} \\
		z &\mapsto z - c_i + h_i e_i \qquad & z &\mapsto z - (1 + h_i\inv)\langle e_i, z \rangle e_i + c_i +  e_i
	\end{align*}
	where $\langle e_i, z \rangle$ is the coefficient of $e_i$ in $z$. 
	Consider the radial neural network with widths $(n, n+1, \dots, n+N, m)$, whose affine transformations and activations are given by:
	\begin{itemize}
		\item 	For $i = 1, \dots, N$ the affine transformation from layer $i-1$ to layer $i$ is given by  $z \mapsto T_i \circ S_{i-1} (z)$, where $S_0 = \id_{\R^n}$.
		\item The activation function at the $i$-th hidden layer is   Step-ReLU on $\R^{n+i}$, that is:
		\begin{align*}
			\rho_i : \R^{n+i} & \longrightarrow \R^{n+i}, \qquad	z \longmapsto 
			\begin{cases} 	z & \qquad \text{\rm if $|z| \geq 1$} \\
				0 & \qquad \text{\rm otherwise} 	
			\end{cases} 	
		\end{align*}
		\item The affine transformation from layer $i=N$ to the output layer is
		\[ z\mapsto \Phi_{L, f,\mathbf{c}} \circ S_N(z) \]
		where $\Phi_{L, f,\mathbf{c}}$ is the affine transformation given by:
		\[
		\Phi_{L, f,\mathbf{c}} : \R^{n+N} \to \R^m, \qquad
		x + \sum_{i=1}^N a_i e_i \  \mapsto \  L(x) + \sum_{i=1}^N a_i(f(c_i) - L(c_i)) \]
		which can be shown to be affine  when $L$ is affine. 
		Indeed, write $L(x) = Ax + b$ where $A$ is a matrix in $\R^{m \times n}$ and $b \in R^{m}$ is a vector. Then 
		$\Phi_{L, f, \bfc}$ is the composition of the linear map given by the matrix
		\[  \begin{bmatrix}
			A & f(c_1) - L(c_1) & f(c_2) - L(c_2)& \cdots & f(c_N) - L(c_N) 
		\end{bmatrix} \in \R^{m \times (n + N)} \]
		and translation by $b \in \R^m$. 
		Note that we regard each $f(c_i) - L(c_i) \in \R^m$ as a column vector in the matrix above. 
	\end{itemize}
	
	
	We claim that the feedforward function of the above radial neural network is exactly $F$. To show this, we first state a lemma, whose (omitted) proof is an elementary computation.
	
	\begin{sublemma}\label{lem:ST}	For $ i= 1, \dots, N$, the composition	$S_i \circ T_i$ is the embedding $\R^{n+i-1} \hookrightarrow \R^{n+i}$. \end{sublemma}
	
	Next, recursively define $G_i : \R^n \to \R^{n+i}$ via 
	\[ G_i = S_i \circ \rho_{i} \circ T_i \circ G_{i-1}, \]
	where $G_0 = \id_{\R^n}$. The function $G_i$ admits an direct formulation:
	
	\begin{subprop}\label{prop:FN} For $i =0, 1, \dots, N$, we have:
		\[
		G_i(x) = \begin{cases}
			x & \qquad \text{\rm if  $x \notin \bigcup_{j=1}^i B_{r_j}(c_j)$}\\
			c_j + e_j & \qquad \text{\rm where $j \leq i$ is the smallest index with $x \in B_{r_j} (c_j)$}
		\end{cases}.
		\]
	\end{subprop}
	
	\begin{proof}
		We proceed by induction.  The base step $i=0$ is immediate. For the induction step, assume the claim is true for $i-1$, where $0 \leq i-1 <N$. There are three cases to consider.
		
		{\bf Case 1}. Suppose $x \notin  \bigcup_{j=1}^{i} B_{r_j}(c_j)$. Then in particular $x \notin \bigcup_{j=1}^{i-1} B_{r_j}(c_j)$, so the induction hypothesis implies that $G_{i-1}(x) = x$. Additionally, $x \notin B_{r_i}(c_i)$, so:
		\[ |T_i(x)| = |x-c_i + h_i e_i|  = \sqrt{ |x-c_i| + h_i^2}  \geq \sqrt{r_i^2 + 1 - r_i^2} = 1. \]
		Using the definition of $\rho_{i}$ and Lemma \ref{lem:ST}, we compute:
		\[ G_i(x) = S_i \circ \rho_{i} \circ T_i \circ G_{i-1}(x) =  S_i \circ \rho_{i} \circ T_i (x)  = S_i \circ T_i(x) = x. \]
		
		{\bf Case 2}. Suppose $x \in B_j \setminus  \bigcup_{k=1}^{j-1} B_{r_k}(c_k)$ for some $j \leq i-1$. Then the induction hypothesis implies that $G_{i-1}(x) = c_j + e_j$.  We compute:
		\[ |T_i(c_j + e_j)| = |c_j + e_j - c_i + h_i e_i|  > |e_j| = 1. \]
		Therefore, 
		\[ G_i(x) =  S_i \circ \rho_{i} \circ T_i (c_j + e_j)  = S_i \circ T_i(c_j + e_j) = c_j + e_j. \]
		
		{\bf Case 3}. Finally, suppose $x \in B_i \setminus  \bigcup_{j=1}^{i-1} B_{r_j}(c_j)$. The induction hypothesis implies that $G_{i-1}(x) = x$. Since $x \in B_{r_i}(c_i)$, we have:
		\[ |T_i(x)| = |x-c_i + h_i e_i|  = \sqrt{ |x-c_i| + h_i^2}  < \sqrt{r_i^2 + 1 - r_i^2} = 1. \]
		Therefore:
		\[ G_i(x) =  S_i \circ \rho_{i} \circ T_i (x)  = S_i (0) = c_i + e_i. \]
		This completes the proof of the proposition.
	\end{proof}
	
	Finally,  we show that the function $F$ defined at the beginning of the proof is the feedforward function of the above radial neural network. The computation is elementary:
	\begin{align*}
		F_\text{\rm feedforward}  &= \Phi_{L,f,\mathbf{c}} \circ S_N \circ \rho_N \circ T_N \circ S_{N-1} \circ \rho_{N-1} \circ T_{N-1} \circ \cdots S_1 \circ \rho_1 \circ T_1 \\ &  = \Phi_{L,f,\mathbf{c}} \circ G_N \\ & = F
	\end{align*}
	where the first equality follows from the definition of the feedforward function, the second from the definition of $G_N$, and the last from  the case $i=N$ of  Proposition \ref{prop:FN} together with the definition of $\Phi_{L, f,\mathbf{c}}$. This completes the proof of the theorem.
\end{proof}

\subsection{Proof of Theorem \ref{thm:UA-n+m+1}: bounded width UA for asymptotically affine functions}\label{app:UA-n+m+1}

We restate and prove Theorem~\ref{thm:UA-n+m+1}, which strengthens Theorem~\ref{thm:UA-asymp-lin} by providing a bounded width radial neural network approximation of any asymptotically affine function.

\thmUAnplusmplusone*

\begin{proof}
	By the hypothesis on $f$, there exists  an affine function $L : \R^n \to \R^m$ and a compact set $K \subset \R^n$ such that $|L(x) - f(x)| < \epsilon$ for all $x \in \R^n \setminus K$.
	Given $\epsilon >0$, let $N= N(f,K,\epsilon)$ and use Lemma~\ref{lem:def-N} to choose  $c_1, \dots, c_N \in K$ and $r_1, \dots, r_N \in (0,1)$ such that the union of the balls $B_{r_i} (c_i)$ covers $K$, and, for all $i$, we have $f(B_{r_i}(c_i)) \subseteq B_{\epsilon}(f(c_i))$. Let $s$ be the minimal non-zero value of $|f(c_i) - f(c_j)|$ for $i,j \in \{1, \dots, N\}$, that is, $s = \min_{i,j, f(c_i) \neq f(c_j) } |f(c_i) - f(c_j)|$.

	Using the decomposition  $\R^{n+m+1} \cong \R^n \times \R^m \times \R$,
	we write elements of $\R^{n+m+1}$ as $(x,y,\theta)$, where $x \in \R^n$, $y \in \R^m$, and $\theta \in \R$.
	For $i = 1, \dots, N$, set:
	\[
	T_i : \R^{n+m+1}  \to \R^{n+m+1}, \qquad
	(x,y,\theta) \mapsto \left(x - (1-\theta) c_i \ , \  y - \theta \frac{f(c_i) - L(0)}{s} \ , \  (1-\theta) h_i \right)
	\]
	where $h_i = \sqrt{1 - r_i^2}$. Note that $T_i$ is an invertible affine transformation, whose inverse is given by:
	\[
	T_i\inv(x,y,\theta) = \left(x + \frac{\theta}{h_i} c_i \ , \ y + \Bigl(1-\frac{\theta}{h_i}\Bigr) \frac{f(c_i) - L(0)}{s} \ , \ 1 - \frac{\theta}{h_i}\right)
	\]
	
	For $i = 1, \dots, N$, define $G_i: \R^n \to \R^{n+m+1}$ via the following recursive definition:
	\[
	G_i = T_i\inv \circ \rho \circ T_i \circ G_{i-1} ,
	\]
	where $G_0(x) = (x,0,0) : {\R^n \hookrightarrow \R^{n+m+1}}$ is the inclusion, and $\rho : \R^{n+m+1} \to \R^{n+m+1}$ is Step-ReLU on $\R^{n+m+1}$. 
	We claim that, for $x \in \R^n$, we have:
	\[
	G_i(x) = \begin{cases}
		(x,0,0) &\qquad \text{\rm if $x \notin \bigcup_{j=1}^i B_{r_j}(c_j)$} \\
		\Bigl(0,\frac{f(c_j)-L(0)}{s},1\Bigr) & \qquad \text{\rm where $j \leq i$ is the smallest index with $x \in B_{r_j} (c_j)$}
	\end{cases}
	\]
	This claim can be verified by a straightforward induction argument, similar to the one given in the proof of Proposition \ref{prop:FN}, and using the following key facts:
	\begin{itemize}
		\item For $x \in \R^n$, $\bigl|T_i\bigl((x,0,0)\bigr)\bigr| = \bigl|(x-c_i,0,h_i)\bigr| < 1$ if and only if $|x-c_i| < r_i$.  
		\item $T_i\inv(0) = \Bigl(0,\frac{f(c_i)-L(0)}{s},1\Bigr)$. 
		\item $T_i\biggl(\Bigl(0,\frac{f(c_j)-L(0)}{s},1\Bigr)\biggr) = \Bigl(0,\frac{f(c_j) - f(c_i)}{s},0\Bigr)$, which, by the choice of $s$, has norm at least $1$ if $f(c_j) \neq f(c_i)$, and is $0$ if $f(c_j) = f(c_i)$. 
	\end{itemize}
	
	Let $\Phi: \R^{n+m+1} \to \R^m$ denote the affine map sending $(x,y,\theta)$ to $L(x) + s y$. It follows that $F = \Phi \circ G_N$ satisfies
	\[
	F(x) = \begin{cases}
		L(x) &\qquad \text{\rm if $x \notin \bigcup_{j=1}^N B_{r_j}(c_j)$} \\
		f(c_j) & \qquad \text{\rm where $j$ is the smallest index with $x \in B_{r_j} (c_j)$}
	\end{cases}
	\]
	By construction, $F$ is the feedforward function of a radial neural network with $N$ hidden layers whose widths are all $n+m+1$.
	Let $x \in \R^n$. If $x \in K$, let $j$ be the smallest index such that $x\in B_{r_j}(c_j)$. Then $F(x) = f(c_j)$, and, by the choice of $r_j$, we have
	$|F(x) - f(x)| = |f(c_j) - f(x) | < \epsilon$.	Otherwise, $x \in \R^n \setminus K$, and $|F(x) - f(x)| = |L(x) - f(x)| < \epsilon$.
\end{proof}

\subsection{Additional result: bound of $\max(n,m) + 1$}\label{app:additional}

We state and prove an additional bounded width result. In contrast to the results above, the theorem below only holds for functions defined on a compact domain, without assumptions about the asymptotic behavior. The proof is an adaptation of the proof of  \autoref{thm:UA-n+m+1}, so we give only a sketch.

\begin{theorem}\label{thm:UA-n+1}
	Let $f : K \to \R^m$ be a continuous function, where $K$ is a compact subset of $\R^n$. For any $\epsilon >0$, there exists $F : \R^n \to \R^m$ such that:
	\begin{enumerate}
		\item $F$ is the feedforward function of a radial neural network with $N(f,K, \epsilon)$ hidden layers whose widths are all $\max(n,m)+1$.
		\item For any $x \in K$, we have $|F(x) - f(x)| < \epsilon$. 
	\end{enumerate}
\end{theorem}

\begin{proof}[Sketch of proof.]
	The construction appearing in the proof of \autoref{thm:UA-n+m+1} with $L \equiv 0$ can be used to produce a radial neural network with $N(f,K, \epsilon)$ hidden layers with widths $n+m+1$ that approximates $f$ on $K$. (Note that the approximation works only on $K$, as $f$ is not defined outside of $K$.)
	All values in the hidden layers are of the form $(x,0,0)$ or $(0,y,1)$. We can therefore replace $(x,y,\theta) \in \R^{n+m+1}$ by $(x+y,\theta) \in \R^{\max(n,m)}\times \R \cong \R^{\max(n,m)+1}$ everywhere, without affecting any statements about the hidden layers.
	In particular, the transformation $T_i$ becomes
	\[
	T_i : \R^{\max(n,m)+1}  \to \R^{\max(n,m)+1}, \qquad
	(x,\theta) \mapsto
	\left(x - (1-\theta) c_i - \theta \frac{f(c_i)}{s} \ , \  (1-\theta) h_i \right).
	\]
	With this change the final affine map $\Phi$ sends $(x,\theta)$ to $sx$.
	From the rest of the proof of \autoref{thm:UA-n+m+1} it follows that the feedforward function $F$ of the radial network satisfies $|F(x) - f(x)| < \epsilon$ for all $x \in K$.
\end{proof}








\subsection{Additional result: bound of $\max(n,m)$}\label{subsec:additional-additional}

In this section, we prove a different version of the result of the previous section. Specifically, we reduce the bound on the widths to $\max(n,m)$ at the cost of using more layers. Again, we focus on functions defined on a compact domain without assumptions about their asymptotic behavior. Recall the notation $M(f,K,\epsilon)$ from \autoref{lem:def-M} and \autoref{def:NM}.

\begin{theorem}\label{thm:UA-bd=n-app}
	Let $f : K \to \R^m$ be a continuous function, where $K$ is a compact subset of $\R^n$ for $n \geq 2$. For any $\epsilon >0$, there exists $F  : \R^n \to \R^m$ such that:
	\begin{enumerate}
		\item $F$ is the feedforward function of a radial neural network with $2M(f,K, \epsilon/2)$ hidden layers whose widths are all $\max(n,m)$.
		\item For any $x \in K$, we have $|F(x) - f(x)| < \epsilon$. 
	\end{enumerate}
\end{theorem}

\widthsublemmas
\begin{proof} 
	We first consider  the proof in the case $n=m$. Set $M = M(f,K,\epsilon)$. As in Lemma~\ref{lem:def-M}, fix $c_1, \dots, c_M \in K$ and $r_1, \dots, r_M \in (0,1)$ such that, first, the union of the balls $B_{r_i} (c_i)$ covers $K$; second, for all $i$, we have $f \left(B_{r_i} (c_i) \right) \subseteq  B_{\epsilon/2}(f(c_i))$; and third, $|c_i  - c_j| \geq r_i$ for $i \neq j$.
	For $i = 1, \dots, M$, set
	\[ T_i : \R^n \to \R^n, \qquad x \mapsto  \frac{x - c_i}{r_i}, \]
	and recursively define $G_i : \R^n \to \R^{n}$ as 
	$G_i = T_i\inv \circ \rho \circ T_{i} \circ G_{i-1},$
	where $G_0 = \id_{\R^n}$ is the identity on $\R^n$ and $\rho : \R^n \to \R^n$ is Step-ReLU. 
	
	\begin{sublemma}\label{lem:Gi}
		For $i =0, 1, \dots, N$, we have:
		\[ G_i(x) = \begin{cases}
			x & \qquad \text{\rm if  $x \notin \bigcup_{j=1}^i B_{r_j}(c_j)$}\\
			c_j & \qquad \text{\rm where $j \leq i$ is the smallest index with $x \in B_{r_j} (c_j)$}.
		\end{cases} \]
	\end{sublemma}
	
	We omit the full proof of Lemma \ref{lem:Gi}, as it is a standard induction argument similar to Proposition \ref{prop:FN}, relying on the following two facts. First, $|T_i(x)| <1$ if and only if $x \in B_{r_i}(c_i)$. Second,  by the choice of $c_i$, we have $|c_i - c_j | \geq r_i$ for all $i\neq j$. This implies that $|T_i(c_j)| \geq 1$ for $i \neq j$. 
	
	Next, perform the following loop over $i = 1, \dots, M$:
	\begin{itemize}
		\item Set $P_{i-1} = \{ c_1, \dots, c_M \} \cup \{d_1, \dots, d_{i-1} \}$
		\item Choose $d_i$ in $B_{\epsilon/2}( f(c_i))$ that is not colinear with any pair of points in $P_{i-1}$. This is where we use the hypothesis that $n \geq 2$. 
		\item  Let $s_i$ be the {minimum} distance between any point on the line through $c_i$ and $d_i$ and any point in $P_{i-1}\setminus \{c_i\}$.
		\item Let $U_i : \R^n \to \R^n$ be the following affine transformation:
		\[ U_i :  \R^n \to \R^n, \qquad x \mapsto  \frac{x-d_i}{s_i} +  \left( \frac{1}{|c_i - d_i|} - \frac{1}{s_i} \right) \frac{ \langle x - d_i , c_i - d_i \rangle}{ |c_i - d_i|^2}(c_i - d_i) \]
		
		\item Define $H_i : \R^n \to \R^n$ recursively as $H_i = U_i\inv \circ \rho  \circ U_i \circ H_{i-1}$, where $H_0 = \id_{\R^n}$.  
		
	\end{itemize}
	
	We note that the transformation $U_i$ can also be written  as $A_i(x - d_i)$ where $A_i $ is the linear map given by $A_i = \frac{1}{s_i}\text{\rm proj}_{\langle c_i-d_i \rangle^\perp}  +  \frac{1}{|c_i - d_i|}\text{\rm proj}_{\langle c_i-d_i \rangle}$, which involves the projections onto the line spanned by $c_i - d_i$ and onto the orthogonal complement of this line.

	\begin{sublemma}\label{lem:Hi}
		For $i,j = 1, \dots, M$, we have:
		\[ H_i(c_j) = \begin{cases}
			d_j & \qquad \text{\rm if $j \le i$} \\
			c_j & \qquad \text{\rm if $j > i$}
		\end{cases} \]
	\end{sublemma}
	
	\begin{proof}
		It is immediate that $U_i(d_i) = 0$ and $|U_i(c_i) |  = 1/2$.  It is also straightforward to show, using the choice of $s_i$, that $|U_i(p)| \geq 1$ for all $p \in P_{i-1} \setminus \{c_i\}$.
		It follows that $U_i\inv \circ \rho \circ U_i$ sends $c_i$ to $d_i$ and fixes all other points in $P_{i-1}$.
	\end{proof}
	
	\begin{sublemma}
		For  $x \in K$, we have $H_M \circ G_M(x) = d_i$ where $i$ is the smallest index with $x \in B_{r_i} (c_i)$
	\end{sublemma}
	
	\begin{proof} Let $x \in K$. By Lemma \ref{lem:Gi}, we have  that $G_M(x) = c_i$ where $i$ is the smallest index with $x \in B_{r_i}(c_i)$. (We use the  fact that the balls $\{B_{r_i}(c_i)\}$ cover $K$.) By Lemma \ref{lem:Hi}, we have that $H_M(c_i) = d_i$ for all $i$. The result follows.
	\end{proof}
	
	Set $F = H_M \circ G_M$. We see that, for $x \in K$:
	\[ |F(x) - f(x) |  = |d_i - f(x)| \leq |d_i - f(c_i)| + |f(c_i) - f(x)| < \epsilon/2 + \epsilon/2 = \epsilon \]
	where $i$ is the smallest index with $x \in B_{r_i}(c_i)$. We show that $F$ is the feedforward function of a radial neural network with $2M$ hidden layers, all of width equal to $n$. Indeed, take the affine transformations and activations as follows:
	\begin{itemize}
		\item 	For $i = 1, \dots, M$ the affine transformation from layer $i-1$ to layer $i$ is given by  $x \mapsto T_i \circ T_{i-1}\inv (x)$, where $T_0 = \id_{\R^n}$.
		
		\item 	For $i = 1, \dots, M$ the affine transformation from layer $M +i-1$ to layer $M+i$ is given by  $x \mapsto U_i \circ U_{i-1}\inv (x)$, where $U_0 = T_N\inv$.
		\item The activation at each hidden layer is Step-ReLU on $\R^n$ that is $\rho(x) = x$ if $|x| \geq 1$ and $0$ otherwise. 
		
		\item Layer $2M+1$ has the affine transformation $U_M\inv$.
		
	\end{itemize}
	It is immediate from definitions that the feedforward function of this network is $F$.

	To conclude the proof, we discuss the cases where $n \neq m$. Suppose $n < m$ so that $\max(n,m) = m$. Then we can regard $K$ as a compact subset of $\R^m$ and apply the above constructions. Suppose $n > m$ so that $\max(n,m) =n$. Let $\inc : \R^m \hookrightarrow \R^n$. Apply the above constructions to the function $\tilde{f} =  \inc  \circ  f : K \to \R^n$. 
\end{proof}

\section{Model compression proofs}\label{app:mod-comp}

The aim of this appendix is to give a proof of Theorem \ref{thm:mod-comp}. In order to do so, we first (1) provide background on a relevant version of the QR decomposition, and (2) establish basic properties of radial rescaling activations.

\subsection{The QR decomposition}\label{subsec:qr}

In this section, we recall the QR decomposition and note several relevant facts.   For integers $n$ and $m$, let $\left( \R^{n \times m} \right)^\text{\rm upper}$ denote the vector space of upper triangular $n$ by $m$ matrices. 

\begin{theorem}[QR Decomposition]\label{prop:QR-gen}  The following  map is surjective:
	\begin{align*}  O(n) \times \left( \R^{n \times m} \right)^\text{\rm upper} &\longrightarrow \R^{n \times m}\\ 
		Q\ , \ R   \quad  &\mapsto  \quad Q  \circ  R 
	\end{align*}
\end{theorem}

In other words, any matrix can be written as the product of an orthogonal matrix and an upper-triangular matrix. When $m \leq n$, the last $n-m$ rows of any matrix in $\left( \R^{n \times m} \right)^\text{\rm upper}$ are zero, and the top $m$ rows form an upper-triangular $m$ by $m$ matrix. These observations lead to the following ``complete'' version of the QR decomposition, which coincides with the above result when $m \geq n$:

\begin{corollary}[Complete QR Decomposition] \label{prop:QR}  The following  map is surjective:
	\begin{align*} \mu:  O(n) \times \left( \R^{k \times m} \right)^\text{\rm upper} &\longrightarrow  \R^{n \times m}\\ 
		Q \ , \ R   \quad  &\mapsto  \quad Q  \ \circ \  \inc \  \circ  \  R 
	\end{align*}
	where $k = \min(n,m)$ and $\inc : \R^{k} \hookrightarrow \R^n$ is the standard inclusion into the first $k$ coordinates. 
\end{corollary}



We make some remarks:
\begin{enumerate}
	\item There are several  algorithms for computing the QR decomposition of a given matrix. One is  Gram--Schmidt orthogonalization, and another is the method of Householder reflections. The latter has computational complexity  $O(n^2m)$ in the case of a $n \times m$ matrix with $n \geq m$.  The package \texttt{numpy} includes a function \href{https://numpy.org/doc/stable/reference/generated/numpy.linalg.qr.html}{\texttt{numpy.linalg.qr}} that computes the QR decomposition of a matrix using Householder reflections. 
	
	\item In each iteration of the loop in Algorithm \ref{alg:QR-mod-comp}, the method \texttt{QR-decomp} with \texttt{mode = `complete'} takes as input a matrix $A_i$ of size $n_i \times (\nred_{i-1} + 1)$, and produces an orthogonal matrix $Q_i \in O(n_i)$ and an upper-triangular matrix $R_i$ of size $\min(n_i, \nred_{i-1} + 1) \times (\nred_{i-1} + 1)$ such that $A_i = Q_i \circ \inc_i \circ R_i$. Note that  $\nred_i = \min(n_i, \nred_{i-1} + 1)$.

	\item The QR decomposition is not unique in general, or, in other words, the map $\mu$ is not injective in general. For example, if $n >m$, 	 each fiber of $\mu$ contains a copy of the orthogonal group  $O(n-m)$. 
	
	\item  The QR decomposition is unique (in a certain sense) for invertible square matrices. To be precise, let $B_n^+$ be the subset of  of $\left( \R^{n \times n} \right)^\text{\rm upper} $ consisting of upper triangular $n$ by $n$ matrices with positive entries along the diagonal. Both $B_n^+$ and $O(n)$ are subgroups of the general linear group $\GL_n(\R)$, and the multiplication map $O(n) \times B_n^+ \to \GL_n(\R)$  is bijective. However, the QR decomposition is not unique for non-invertible square matrices.  	
\end{enumerate}

\subsection{Radial rescaling functions}\label{subsec:radial-properties}

We now prove the following basic facts about radial rescaling functions:

\begin{lemma}\label{lem:radial-basics} Let $\rho = h^{(n)} : \R^n \to \R^n$ be a radial rescaling function on $\R^n$. 
	\begin{enumerate}
		\item The function $\rho$ commutes with any orthogonal transformation of $\R^n$. That is,
		$\rho \circ Q = Q \circ \rho$  for any $Q \in O(n)$.
		
		\item If $m \leq n$ and  $\inc : \R^m \hookrightarrow \R^n$ is the standard inclusion into the first $m$ coordinates, then:
		$h^{(n)}   \circ \inc = \inc \circ h^{(m)}.$
	\end{enumerate}
\end{lemma}

\begin{proof}  Suppose $Q\in O(n)$ is an orthogonal transformation of $\R^n$. Since $Q$ is norm-preserving, we have $|Q v| =  |v|$ for any $v \in \R^n$. Since $Q$ is linear, we have $Q(\lambda v ) = \lambda Qv$ for any $\lambda \in \R$ and $v \in \R^n$. Using the definition of $a= h^{(n)}$ we compute:
	\begin{align*}
		\rho(Qv) & =  \frac{h(|Qv|)}{|Qv|} Qv =  \frac{h(|v|)}{|v|} Qv = Q\left(  \frac{h(|v|)}{|v|}  v\right) = Q(\rho(v)).
	\end{align*}
	The first claim follows. The second claim is an elementary verification. \end{proof}

More generally, the restriction of the radial rescaling function $\rho$ to a linear subspace of $\R^n$ is a radial rescaling function on that subspace. Given a tuple radial rescaling functions $\boldrho = \left(\rho_i : \R^{n_i} \to \R^{n_i}\right)_{i=1}^L$ suited to widths $\bfn = (n_i)_{i=1}^L$, we write $\brhored = \left(\rhored_i : \R^{\nred_i} \to \R^{\nred_i}\right)$ for the tuple of restrictions suited to the reduced widths $\bfnred$, so that $\rhored_i = \rho_i \biggr\vert_{\R^{\nred_i}}$.

\subsection{Proof of Theorem \ref{thm:mod-comp}}\label{subsec:proof-mod-comp}

Adopting notation from above and Section \ref{sec:mod-comp}, we now restate and prove Theorem \ref{thm:mod-comp}. 

\thmModelCompression*


\begin{proof}
Let $(\bfWred, \bfbred , \bfQ) = 	\text{\rm \texttt{QR-Compress}($\bfW, \bfb$)}$ be the output of Algorithm \ref{alg:QR-mod-comp}, so that $\bfQ \in O(\bfn^\text{\rm hid})$ and  $(\bfWred, \bfbred, \boldrho^\text{\rm red})$ is a neural network with widths $\nred$ and radial rescaling activations $\rho_i^\text{\rm red} = \rho_i \biggr\vert_{\R^{\nred_i}}$.
Let $F = F_{(\bfW, \bfb, \boldrho)}$ denote the feedforward function of the radial neural network with parameters $(\bfW, \bfb)$ and activations $\boldrho$. Similarly, let  $F^\text{\rm red} = F_{(\bfWred, \bfbred, \brhored)}$ denote the feedforward function of the radial neural network with parameters $(\bfWred, \bfbred)$ and activations $\brhored$. Additionally, we have the partial feedforward functions $F_i$ and $\Fred_i$. We show by induction that 
\[ F_i = Q_i \circ \inc_i \circ \Fred_i \]
for any $i = 0,1, \dots, N$.  (Continuing conventions from Sections \ref{subsec:paramspace} and \ref{subsec:mod-comp}, we set $Q_0 = \id_{\R^{n_0}}$, $Q_L = \id_{\R^{n_L}}$, and $\inc_i : \R^{\nred_i} \to \R^{n_i}$ to be the inclusion map.) The base step $i=0$ immediate. For the induction step, let $x \in \R^{n_0}$. Then:
\begin{align*}
	F_i (x) & = \rho_i \left( W_i \circ F_{i-1} (x) + b_i \right) \\ 
	&= \rho_i \left( W_i \circ Q_{i-1} \circ \inc_{i-1} \circ \Fred_{i-1} (x) + b_i \right) \\
	&= \rho_i \left( \begin{bmatrix}
		b_i & W_i \circ Q_{i-1} \circ \inc_{i-1}
	\end{bmatrix} \begin{bmatrix}
		1 \\ \Fred_{i-1}(x)
	\end{bmatrix} \right) \\
	&= \rho_i \left( Q_i \circ \inc_i \circ \begin{bmatrix}
		b_i^\text{\rm red} & W_i^\text{\rm red} 
	\end{bmatrix} \begin{bmatrix}
		1 \\ \Fred_{i-1}(x)
	\end{bmatrix} \right) \\
	& = Q_i \circ \inc_i \circ \rho_i \biggr\vert_{\R^{\nred_i}} \left( W_i^\text{\rm red}  \circ \Fred_{i-1}(x) + b_i^\text{\rm red} \right) \\
	& = Q_i \circ \inc_i \circ \Fred_i
\end{align*}
The first equality relies on the definition of the partial feedforward function $F_i$; the second on the induction hypothesis; the fourth on an inspection of Algorithm \ref{alg:QR-mod-comp}, noting that $R_i = [b_i^\text{\rm red} \  W_i^\text{\rm red} ]$; the fifth on the results of Lemma \ref{lem:radial-basics}, observing that $\rho_i \circ \inc_i = \rho_i \vert_{\R^{\nred_i}}= \inc_i \circ \rho_i^\text{\rm red}$; and the sixth on the definition of $\Fred_{i}$. In the case $i=L$, we have:
\[ F = F_L  = Q_L \circ \inc_L \circ \Fred_L = \Fred \]
since $Q_L = \inc_L = \id_{\R^{n_L}}$ and $\Fred_L = \Fred$. The theorem now follows.
\end{proof}

The techniques of the above proof can be used to show that the action of the group $O(\bfn^\text{\rm hid})$ of orthogonal change-of-basis symmetries on the parameter space $\Par(\bfn)$ leaves the feedforward function unchanged.  We do not use this result directly, but state is precisely it nonetheless:

\begin{proposition} \label{prop:rnns-invariance}
Let $(\bfW, \bfb, \boldrho)$ be a radial neural network with widths  vector $\bfn$. Suppose $\bfg \in O(\bfn^\text{\rm hid})$. Then the original and transformed networks have the same feedforward function:
\[ F_{(\bfg \cdot \bfW , \  \bfg \cdot \bfb  ,  \ \boldrho)} = F_{(\bfW  , \  \bfb  ,  \ \boldrho )} \]
\end{proposition}

In other words, fix parameters $(\bfW, \bfb) \in \Par(\bfn)$, radial rescaling activations $\boldrho$, and $\bfg \in O(\bfn^\text{\rm hid})$. Then  the radial neural network with parameters $(\bfW, \bfb)$ has the same feedforward function as the radial neural network with transformed parameters $(\bfg \cdot \bfW, \bfg \cdot \bfb)$, where we take radial rescaling activations $\boldrho$ in both cases.

We remark that Proposition \ref{prop:rnns-invariance} is analogous to the ``non-negative homogeneity'' (or ``positive scaling invariance'') of the pointwise ReLU activation function\footnote{See  Armenta and Jodoin, {\it The Representation Theory of Neural Networks},
arXiv:2007.12213;  Dinh,  Pascanu,  Bengio, and  Bengio, {\it Sharp Minima Can Generalize For Deep Nets}, ICML 2017;   Meng,  Zheng,  Zhang,  Chen,  Ye, Ma,  Yu, and Liu, {\it G-SGD: Optimizing ReLU Neural Networks in its Positively Scale-Invariant Space}, 2019; and  Neyshabur,  Salakhutdinov, and  Srebro. {\it Path-SGD: path-normalized optimization in deep neural networks}, NIPS’15.}. In that setting, instead of considering the product of orthogonal groups $O(\bfn^\text{\rm hid})$, one considers the rescaling action of the following subgroup of $\prod_{i=1}^{L-1} \GL_{n_i}$:
$$G = \left\{ \bfg = (g_i) \in \prod_{i=1}^{L-1} \GL_{n_i} \ | \ \text{ each  $g_i$ is diagonal with positive diagonal entries}\right\}$$
Note that $G$ is isomorphic to the product  $\prod_{i=1}^{L-1} \R_{>0}^{n_i}$, and the action on $\Par(\bfn)$ is given by the same formulas as those appearing near the end of Section \ref{subsec:paramspace}. 
The feedforward function of a MLP with pointwise ReLU activations is invariant for the action of $G$ on $\Par(\bfn)$.

\section{ Projected gradient descent proofs}\label{app:proj-gd}

In this section, we give a proof of Theorem \ref{thm:proj-gd}, which relates projected gradient descent for a representation with dimension $\bfn$ to (usual) gradient descent for the  corresponding reduced  representation with dimension vector  $\bfnred$. This proof requires some set up and background resutls. 

\subsection{Gradient descent and orthogonal symmetries}\label{app:gd-orth}

We first prove a result that gradient descent commutes with invariant orthogonal transformations. This section is general and departs from the specific case of radial neural networks. 

\renewcommand{\L}{\mathcal{L}}

\subsubsection{Setting}
Let $\L : V = \R^p \to \R$ be a smooth function. Semantically, $V$ is a the parameter space of a neural network and $\L$ the loss function with respect to a batch of training data.  The differential $d\L_v$ of $\L$ at $v \in V$ is  row vector, while the gradient $\nabla_v \L$ of $\L$ at $v$ is a column vector\footnote{Following usual conventions, we regard column vectors as elements of $V$ and  row vectors as elements of the dual vector space $V^*$. The differential $d\L_v$ of $\L$ at $v \in V$ is also known as the Jacobian of $\L$ at $v \in V$.}:
\begin{align*}  d \mathcal L_v = \begin{bmatrix}
	\frac{\partial \L}{\partial x_1} \bigg\vert_v   & \cdots & \frac{\partial  \L}{\partial x_p} \bigg\vert_v 	\end{bmatrix}  \qquad \qquad \nabla_v \L = \begin{bmatrix}
	\frac{\partial \L}{\partial x_1} \bigg\vert_v \\   \vdots \\    \frac{\partial  \L}{\partial x_p} \bigg\vert_v 
\end{bmatrix}	
\end{align*}
Hence $\nabla_v \L$ is the transpose of $d\L_v$, that is: $\nabla_v \L = (d \L_v)^T$.  A step of gradient descent with respect to $\L$ at learning rate $\eta >0$ is defined as:
\begin{align*}
\gamma = \gamma_\eta : V & \longrightarrow V\\
v &\longmapsto v - \eta \nabla_v \L
\end{align*}
We drop $\eta$ from the notation when it is clear from context. For any $k \geq 0$, we denote by $\gamma^k$ the $k$-fold composition of the gradient descent map $\gamma$:
\[ \gamma^k = \overbrace{\gamma \circ \gamma \circ \cdots \circ \gamma}^k \]

\subsubsection{Invariant group action}

Now suppose $\rho : G \to \GL(V)$ is an action of a Lie group $G$ on $V$ such that $\L$ is $G$-invariant, i.e.:
\[ \L(\rho(g)(v)) = \L(v)\]
for all $g \in G$ and $v \in V$. We write simply $ g\cdot v$ for $\rho(g)(v)$, and $g$ for $\rho(g)$. 

\begin{lemma}\label{lem:nabla-g} For any $v \in V$ and $g \in G$, we have:
$$\nabla_{v} \L =g^T \cdot ( \nabla_{g \cdot v} \L) $$ \end{lemma}

\begin{proof} The proof is a computation:
\begin{align*}
	\nabla_{v} \L &= (d_{v} \L)^T   =  ( d (\L \circ g)_v)^T = (d\L_{g \cdot v}  \circ d g_v)^T = (d \L_{g \cdot v} \circ  g)^T=  g^T \cdot  ( d \L_{g \cdot v})^T \\ &= g^T \cdot ( \nabla\L_{g \cdot v}) 
\end{align*}
The second equality relies on the hypothesis that  $\L \circ g = \L$, the third on the chain rule, and the fourth on the fact that  $dg_v = g$ since $g$ is a linear map. 
\end{proof}

One can perform the computation of the proof in coordinates, for $i = 1, \dots, p$:
\begin{align*}
\left( \nabla_v \L \right)_i &=  \left( d \L_v \right)^i = \frac{\partial \L}{\partial x_i} \biggr \vert_v = \frac{\partial (\L\circ g)}{\partial x_i} \biggr \vert_v  = \frac{\partial \L}{\partial x_j} \biggr \vert_{gv} \frac{\partial g_j}{\partial x_i} \biggr \vert_v \\
&= \left( \nabla_{gv} \L \right)_j g_j^i  = (g^T)^j_i  \left( \nabla_{gv} \L \right)_j = \left( g^T \cdot \nabla_{gv} \L \right)_i
\end{align*}

\subsubsection{Orthogonal case}

\begin{figure}[t]
\centering
\includegraphics[width=0.49\textwidth,trim=315 200 0 0, clip]{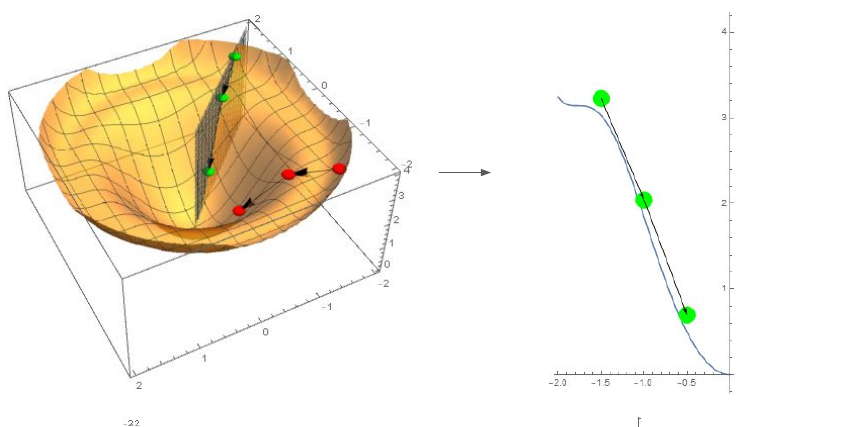}
\caption{Illustration of Lemma \ref{lem:orth-gd}. If the loss is invariant with respect to an orthogonal transformation $Q$ of the parameter space, then optimization of the network by gradient descent is also invariant with respect to $Q$. (Note: in this example, projected and usual gradient descent match; this is not the case in higher dimensions, as explained in  \ref{appsubsec:ex-131}.) }
\label{fig:orth-gd}
\end{figure}

Furthermore, suppose the action of $G$ is by orthogonal transformations, so that $\rho(g)^T = \rho(g)\inv$ for all $g \in G$. Then Lemma \ref{lem:nabla-g} implies that
\begin{equation}\label{eq:g-nabla-orth}
\nabla_{g \cdot v} \L = g \cdot \nabla_{v} \L
\end{equation} 
for any $v \in V$ and $g \in G$. The proof of the following lemma is immediate from Equation \ref{eq:g-nabla-orth}, together with the definition of $\gamma$. See Figure \ref{fig:orth-gd} for an illustration. 

\begin{lemma}\label{lem:orth-gd} Suppose the action of $G$ on $V$ is by orthogonal transformations, and that $\L$ is $G$-invariant. Then the action of $G$ commutes with gradient descent (for any learning rate). That is, 
\[ \gamma^k(g \cdot v) = g \cdot \gamma^k(v)\]
for any $v \in V$, $g \in G$, and $k \geq 0$. 
\end{lemma}

\subsection{Gradient descent notation and set-up}\label{app:set-up}

We now turn our attention back to radial neural networks. In this section, we recall notation from above, and introduce new notation that will be relevant for the formulation and proof of Theorem \ref{thm:proj-gd}.

\subsubsection{Merging widths and biases}
Let $\bfn = (n_0, n_1, n_2, \dots, n_{L-1}, n_L)$ be the widths vector of an MLP. Recall the definition of $\Par(\bfn)$ as the parameter space of all possible choices of trainable parameters:
\[ \Par(\bfn) = \left( \R^{n_1 \times n_0} \times \R^{n_2 \times n_1} \times \cdots \times \R^{n_L \times n_{L-1}} \right)  \times \left(\R^{n_1}  \times \R^{n_2} \times  \cdots \times \R^{n_L} \right) \]
We have been denoting an element therein as a pair  of tuples $(\bfW, \bfb)$ where $\bfW = (W_i \in \R^{n_i \times n_{i-1}})_{i=1}^L$ are the weights and $\bfb = (b_i \in \R^{n_i})_{i=1}^L$ are the biases. 
However, in this appendix we adopt different notation. Observe that, placing each bias vector as a extra column on the left of the weight matrix, we obtain matrices:
\[ A_i = [b_i \ W_i]  \ \in   \  \R^{n_i \times (1+ n_{i-1})}. \]
Thus, there is an isomorphism:
\[ \Parn \simeq  \bigoplus_{i=1}^L \R^{n_i \times (n_{i-1} + 1)} = \R^{n_1 \times (n_0 + 1)} \times \R^{n_2 \times (n_1 + 1)} \times \cdots \times \R^{n_L \times (n_{L-1} + 1)} \]
In this appendix, we regard an element of $\Parn$ as a tuple of `merged' matrices $\bfA = (A_i \in \R^{n_i \times (1+ n_{i-1})})_{i=1}^L$. We now define convenient maps to translate between the merged notation and the split notation.
For each $i$, define the extension-by-one map from $\R^{n_i}$ to $\R \times \R^{n_i} \simeq \R^{n_i+1}$ as follows:
\begin{equation}\label{eq:ext-map}
\ext_i : \R^{n_i} \to \R^{n_i+1}\qquad v =(v_1, v_2,\dots, v_{n_i}) \mapsto (1, v_1, v_2, \dots, v_{n_i})
\end{equation}
Observe that, for any $i$ and $x \in \R^{n_{i-1}}$, we have $$A_i \circ \ext_{i-1} (x) = W_i x + b_i.$$  Consequently, the $i$-th partial feedforward function can be defined recursively as:
\begin{equation}
\label{eq:feedforward-merged}
F_i = \rho_i \circ A_i \circ \ext_{i-1} \circ F_{i-1}
\end{equation}
where $\rho_i : \R^{n_i}  \to \R^{n_i}$ is the activation\footnote{In this general formulation, $\rho_i$ can be any piece-wise differentiable function; for most of the rest of the paper we will be interested in the case where $\rho_i$ is a radial rescaling function.} at the $i$-th layer, and $F_0$ is the identity on $\R^{n_0}$.


\subsubsection{Orthogonal change-of-basis action}
To describe the orthogonal change-of-basis symmetries of the parameter space in the merged notation, recall the following product of orthogonal groups, with sizes  corresponding to the widths of the hidden layers:
\[ O(\bfn^\text{\rm hid}) = O(n_1) \times O(n_2) \times \cdots \times O(n_{L-1}) \]
In the merged notation, the element $\mathbf{Q} = (Q_\inx)_{\inx =1}^{L-1} \in O(\mathbf{n}^\text{\rm hid})$  transforms  $\bfA \in \Par(\bfn)$ as:
\begin{equation}\label{eq:O-hidden-action-neur}
\bfA \quad  \mapsto \quad \mathbf{Q} \cdot  \bfA :=   \left( Q_\inx \circ A_\inx \circ  \begin{bmatrix} 1 & 0 \\ 0 & Q_{\inx-1}\inv \end{bmatrix}  \right)_{\inx=1}^L
\end{equation}  
where $Q_0 = \id_{n_0}$ and $Q_L = \id_{n_L}$. 

\subsubsection{Model compression algorithm}
We now restate Algorithm \ref{alg:QR-mod-comp} in the merged notation. We emphasize that Algorithms \ref{alg:QR-mod-comp} and \ref{alg:QR-mod-comp-merged} are mathematically equivalent; the later simply uses more compact notation.

\begin{algorithm}\label{alg:QR-mod-comp-merged}
\SetKwFunction{QRdecompCom}{QR-decomp}
\SetKwFunction{QRdecompRed}{QR-decomp}
\SetKwInOut{Input}{input}
\SetKwInOut{Output}{output}
\SetKwInOut{Initialize}{initialize}
\DontPrintSemicolon
\Input{$\bfA \in \Par(\bfn)$}
\Output{$\mathbf{Q} \in O(\bfn^\text{\rm hidden})$ and $\mathbf{V} \in \Par(\bfnred)$}
\BlankLine

$\bfQ, \bfV \gets [\ ], [\ ]$  
\tcp*[r]{initialize output matrix lists}

$M_1 \gets A_1$

\For(\tcp*[r]{iterate through layers \vspace{-\baselineskip}})
{$ \inx \leftarrow 1$ \KwTo $L-1$   }{ 
		$Q_\inx, R_\inx \gets $ \QRdecompCom{$M_\inx$, \ \texttt{mode = `complete'} } \tcp*[r]{$M_\inx = Q_\inx \circ \inc_\inx \circ R_\inx$}
	Append $Q_\inx$ to $\bfQ$\;
	Append $R_\inx$ to $\bfV$
	\tcp*[r]{reduced merged weights for layer $i$}
	
	Set $M_{\inx+1} \gets  A_{\inx+1}\circ \begin{bmatrix} 1 & 0 \\ 0 & Q_\inx \circ \inc_i  \end{bmatrix} \qquad \qquad$ \tcp*[r]{transform next layer} 
	
}
Append $M_L$ to $\bfV$\;
\BlankLine
\KwRet $\mathbf{Q}$, $\mathbf{V}$
\caption{QR Model Compression (\texttt{QR-compress})}
\end{algorithm}

We explain the notation. As noted in Appendix \ref{app:UA-notation}, the symbol `$\circ$' denotes composition of maps, or  matrix multiplication in the case of linear maps. 
The standard inclusion  ${\inc}_i  : \R^{\nred_{i}} \hookrightarrow \R^{n_i}$ maps into the first  $\nred_\inx$ coordinates. 
As a matrix, $\Inc_i \in \R^{n_i \times \nred_i}$ has ones along the main diagonal and zeros elsewhere. 
The method \texttt{QR-decomp} with \texttt{mode = `complete'} computes the complete QR decomposition of the $n_\inx \times (1 +  \nred_{\inx-1})$ matrix $M_\inx$ as  $Q_\inx \circ \inc_i \circ  R_\inx$ where $Q_\inx \in O(n_\inx)$ and $R_\inx$ is upper-triangular of size $\nred_ \inx \times (1+\nred_{i-1})$. The definition of $\nred_\inx$  implies that either  $\nred_\inx = \nred_{\inx-1} + 1$  or $\nred_\inx = n_\inx$.  The matrix $R_\inx$ is of size $\nred_ \inx \times \nred_{\inx}$ in the former case and of size $n_\inx \times  (1 +  \nred_{\inx-1})$ in the latter case.

\subsubsection{Gradient descent definitions}

As in Section \ref{subsec:proj-gd}, we fix: 

\begin{itemize}
\item a widths vector $\bfn = (n_0, n_1, \dots, n_L)$.
\item a tuple $\boldrho = (\rho_1, \dots, \rho_L)$ of radial rescaling activations, where $\rho_i : \R^{n_i} \to \R^{n_i}$ for $\inx = 1, \dots, L$. 

\item a batch of training data  $\{ (x_j, y_j)\} \subseteq \R^{n_0} \times \R^{n_L} =  \R^{\nred_0} \times \R^{\nred_L} $.

\item a cost function $\mathcal C : \R^{n_L} \times \R^{n_L} \to \R$
\end{itemize}

As a result, we have a loss function on $\Par(\bfn)$:
\[
\mathcal L:  \Par(\bfn)\rightarrow \R \qquad \qquad
\mathcal L( \bfA ) = \sum \mathcal C( F_{(\bfA, \boldrho)} ( x_j),  y_j)
\]
where  $F_{(\bfA, \boldrho)}$ is the feedforward of the radial neural network with (merged) parameters $\bfA$ and activations $\boldrho$.  We emphasize that the loss function $\mathcal L$ depends on the batch of training data chosen above; however, for clarity, we omit extra notation indicating this dependency since the batch of training data is fixed throughout this discussion. Similarly, we have:
\begin{itemize}
\item the reduced widths vector $\bfnred = (\nred_0, \nred_1, \dots,\nred_L)$. 
\item the restrictions $\brhored = (\rhored_1, \dots, \rhored_L)$, where $\rhored_i : \R^{\nred_i} \to \R^{\nred_i}$ for $\inx = 1, \dots, L$. 
\end{itemize}
Using the fact that $\nred_0 = n_0$ and $\nred_L = n_L$, there is a loss function  on $\Parnred$:
\[
\mathcal L_\text{\rm red} :  \Parnred \rightarrow \R \qquad \qquad
\mathcal L_\text{\rm red}( \bfB  ) = \sum \mathcal C(F_{(\bfB, \brhored)} ( x_j),  y_j)
\]
where  $F_{(\bfB, \brhored)}$ is the feedforward of the radial neural network with parameters $\bfB \in \Parnred$ and activations $\brhored$. (Again, technically speaking, the loss function $\mathcal L_\text{\rm red}$ depends on the batch of training data fixed above.) For any learning rate $\eta >0$,  we obtain a  gradient descent maps:
\begin{align*}
\gamma : \Par(\bfn) & \to \Par(\bfn)  \qquad &	\gamma_{\text{\rm red}} : \Par(\bfnred) &\to \Par(\bfnred)    \\
\bfA &\mapsto \bfA  -  \eta \nabla_{\bfA} \mathcal L  \qquad & \bfB &\mapsto \bfB -  \eta \nabla_{\bfB} \mathcal L_\text{\rm red}
\end{align*}

\subsection{The interpolating space}\label{appsubsec:interpolating}

In this section, we introduce a subspace $\Parintn$ of $\Parn$, that, as we will later see, interpolates between $\Parn$ and $\Parnred$. 

Let $\Parintn $ denote the subspace of $\Parn$ consisting of those $\mathbf{T} = (T_1, \dots, T_L) \in  \Parn$   for which the bottom left $(n_\inx - \nred_\inx) \times (1 + \nred_{\inx -1})$ block of $T_\inx$ is zero for each $\inx$. Schematically:
\[ T_\inx = \begin{bmatrix}
* & * \\
0 & *
\end{bmatrix} \]
where the rows are divided as $\nred_i$ on top and $n_i - \nred_i$ on the bottom, while the columns are divided as $(1  + \nred_{i-1})$ on the left and $n_{i-1} - \nred_{i-1}$ on the right.  Let 
\[ \iota_1 : \Parintn \hookrightarrow \Parn \]
be the inclusion. The following proposition follows from an elementary  analysis of the workings of Algorithm~\ref{alg:QR-mod-comp-merged} (or, equivalently, Algorithm \ref{alg:QR-mod-comp}). 


\begin{proposition}\label{prop:qinva-in-int}
Let $\bfA \in \Par(\bfn)$ and let $\bfQ \in O(\bfn^\text{\rm hid})$  be the tuple of orthogonal matrices produced by  Algorithm  \ref{alg:QR-mod-comp-merged}. Then $\bfQ\inv \cdot \bfA$ belongs to $\Parintn$. 
\end{proposition}


Define a map
\[ q_1  : \Par(\bfn)  \to \Par^{\text{\rm int}} ( \bfn) \]
by taking $\mathbf{A} \in \Par(\bfn)$ and zeroing out the bottom left $(n_\inx - \nred_\inx) \times (1 + \nred_{\inx -1})$ block of $A_\inx$ for each $i$. Schematically:
\[
\bfA = \left( A_i = \begin{bmatrix}
* & * \\
* & *
\end{bmatrix}  \right)_{i=1}^L  \  \mapsto  \  q_1(\bfA) = \left( \begin{bmatrix}
* & * \\
0 & *
\end{bmatrix} \right)_{i=1}^L
\]
It is straightforward to check that $q_1$ is a well-defined, surjective linear map.  The transpose of $q_1$ is the inclusion $\iota_1$. We summarize the situation in the following diagram:
\begin{equation}\label{diag:int}
\xymatrix{\Par^{\text{\rm int}} ( \bfn) \ar@/^/[rr]^{\iota_1}  & & \Par ( \bfn) \ar@/^/[ll]^{q_1} 
} \end{equation}
We observe that the composition $q_1 \circ \iota$ is the identity on $\Parintn$.

\subsection{Projected gradient descent and model compression}\label{appsubsec:projGD-QR}


Recall from Section \ref{subsec:proj-gd} that the {\it  projected gradient descent} map on $\Par(\bfn)$ is given by:
\begin{align*}
\gamma_{\text{\rm proj}} : \Par(\bfn) \to \Par(\bfn)    ,\qquad 
\bfA  \mapsto \Proj \left( \bfA  -   \eta  \nabla_{\bfA} \mathcal L  \right)
\end{align*}
where $\bfA = (\bfW, \bfb)$ are the merged parameters (Appendix \ref{app:set-up}), and, in the notation of the previous section, the map $\Proj$ is $\iota_1 \circ q_1$. 
To reiterate, while all entries of each weight matrix and each bias vector contribute to the computation of the gradient $\nabla_{\bfA } \mathcal L   = \nabla_{(\bfW, \bfb) } \mathcal L$, only those not in the bottom left submatrix get updated under the projected gradient descent map $ \gamma_\text{\rm proj}$. 





Let $\bfV, \bfQ = \texttt{QR-Compress}(\bfA)$ be the outputs of Algorithm \ref{alg:QR-mod-comp-merged} (which is equivalent to Algorithm \ref{alg:QR-mod-comp}), so that $\bfV = (\bfWred, \bfbred)\in \Par(\bfnred)$ are the parameters of the compressed model corresponding to the full model with merged parameters $\bfA= (\bfW, \bfb)$, and $\bfQ\in O(\bfn^\text{\rm hid})$ is an orthogonal change-of-basis symmetry of the parameter space.
Moreover, set $\bfT = \bfQ\inv \cdot \bfA  \in \Parintn$, where we use the change-of-basis action from Appendix \ref{app:set-up} and Proposition \ref{prop:qinva-in-int}. We have the following rephrasing of Theorem \ref{thm:proj-gd}.

\begin{theorem}[Theorem \ref{thm:proj-gd}]
\label{thm:proj-gd-ap}
Let $\bfA \in \Par(\bfn)$, and let $\bfV, \bfQ, \bfT$ be as above. For any $k \geq 0$:
\begin{enumerate}
	\item $\gamma^k( \bfA) = \bfQ \cdot \gamma^k(\bfT)$ 
	\item $\gamma_\text{\rm proj}^k (\bfT) =   \gamma_{\text{\rm red}}^k (\bfV)   + \bfT - \bfV.$
\end{enumerate}
\end{theorem} 

More precisely, the second equality  is  $\gamma_\text{\rm proj}^{k}( \bfT ) =   \iota(	\gamma_\text{\rm red}^k (\bfV ) ) + \bfT - \iota(\bfV)$ where $\iota : \Par(\bfnred) \hookrightarrow \Par(\bfn)$ is the inclusion into the top left corner in each coordinate. Also, in the statement of Theorem \ref{thm:proj-gd}, we have $\bfU = \bfT - \bfV$.

We summarize this result in the following diagram. The left horizontal maps indicate the addition of $\bfU = \bfT - \bfV$, the right horizontal arrows indicate the action of $\bfQ$, and the vertical maps are various versions of gradient descent. The shaded regions indicate the (smallest) vector space to which the various  representations naturally belong. 


\[
\begin{tikzpicture}[scale=0.92]
\matrix (m) [matrix of math nodes, row sep=7em,
column sep=2.1em]{ 
	\bfV & & &  \bfT & & & \bfW \\
	\gamma_{\text{\rm red}}^k (\bfV)    & &  \gamma_\text{\rm proj}^k (\bfT) &  &  \gamma^k(\bfT)   &  &  \gamma^k( \bfW) \\};
\path[|->]
(m-1-1) edge node [above] {\scriptsize $+ \bfT - \bfV$} (m-1-4)
(m-1-4) edge node [left]  {\scriptsize {proj-GD on $\Par(\bfn)$}} (m-2-3)
(m-2-1) edge node [below] {\scriptsize $+ \bfT-\bfV$} (m-2-3)
(m-1-1) edge node [left] {\scriptsize GD on $\Par(\bfnred)$}  (m-2-1)
(m-1-4) edge node [right] {\scriptsize GD on $\Par(\bfn)$} (m-2-5)
(m-2-5) edge node [below] {\scriptsize $\bfQ \cdot$} (m-2-7)
(m-1-4) edge node [above] {\scriptsize $\bfQ \cdot$} (m-1-7)
(m-1-7) edge node [right] {\scriptsize GD on $\Par(\bfn)$} (m-2-7);
\draw [dashed,opacity=0.5,fill=red,fill opacity=0.02, smooth,samples=50] 
(-7.5,3) -- (-1.3,3) -- (-3.9,-2.5) -- (-7.5,-2.5) -- cycle;
\draw [dashed,opacity=0.5,fill=green,fill opacity=0.02, smooth,samples=50] 
(-1.3,3) -- (2.4,3) -- (-0.2,-2.5) -- (-3.9,-2.5);
\draw [dashed,opacity=0.5,fill=blue,fill opacity=0.02, smooth,samples=50] 
(2.4,3) -- (7.5,3) -- (7.5,-2.5) -- (-0.2,-2.5);
\node at (-4.6,2.6) {\underline{\large $ \Par ( \bfnred)$}};
\node at (0,2.6) {\underline{\large $ \Par^{\text{\rm int}}(\bfn) $}};
\node at (4.5,2.6) {\underline{\large $ \Par(\bfn)$}};

\end{tikzpicture}
\]

\subsection{Proof of Theorem \ref{thm:proj-gd}}\label{appsubsec:proof-projGD}

We begin by explaining the sense in which $ \Par^{\text{\rm int}} ( \bfn) $ interpolates between $\Par(\bfn)$ and $\Par(\bfnred)$. One extends Diagram \ref{diag:int} as follows:
\[
\xymatrix{ \Par(\bfnred)  \ar@/^/[rr]^{\iota_2} & & \Par^{\text{\rm int}} ( \bfn) \ar@/^/[ll]^{q_2}  \ar@/^/[rr]^{\iota_1}  & & \Par ( \bfn) \ar@/^/[ll]^{q_1} 
}
\]
\begin{itemize}
\setlength\itemsep{10pt}
\item The map \[ \iota_2 : \Par(\bfnred) \hookrightarrow  \Par^{\text{\rm int}} ( \bfn) \] takes $\bfB = (B_i) \in \Par(\bfnred)$ and pad each matrix with $n_i - \nred_i$ rows of zeros on the bottom and $n_{i-1}- \nred_{i-1}$ columns of zeros on the right:
\[ \bfB = \left( B_i \right)_{i=1}^L  \  \mapsto  \  \iota_2(\bfB) = \left( \begin{bmatrix}
	B_i & 0 \\
	0 & 0
\end{bmatrix} \right)_{i=1}^L \]
It is straightforward to check that $\iota_2$ is a well-defined   injective linear map.   

\item The map \[ q_2 : \Par^{\text{\rm int}} ( \bfn)  \rightarrow \Par(\bfnred) \] extracts from  $\mathbf{T}$ the top left $\nred_i \times (1 +\nred_{i-1})$ matrix:
\[ \bfT = \left( T_i  =  \begin{bmatrix}
	T_i^{(1)} & T_i^{(2)} \\
	0 & T_i^{(4)}
\end{bmatrix}\right)_{i=1}^L  \  \mapsto  \  q_2(\bfT) = \left( T_i^{(1)} \right)_{i=1}^L \]
It is straightforward to check  that $q_2$ is a surjective linear map. The transpose of $q_2$ is the inclusion $\iota_2$. \\
\end{itemize}

\begin{lemma}\label{lem:repint} We have the following:
\begin{enumerate}
	\item The inclusion  $\iota : \Par( \bfnred) \hookrightarrow \Par(\bfn)$ coincides with the composition $\iota_1 \circ \iota_2$, and  commutes with the loss functions:
	\[
	\xymatrix{
		\Par( \bfnred) \ar@{^{(}->}[rr]^-{\iota_1 \circ \iota_2 = \iota} \ar[dr]_{\mathcal L_\text{\rm red}} & &  \Par(\bfn) \ar[ld]^{\mathcal L} \\
		& \R& 
	}\]
	\item The following diagram commutes:
	\[\xymatrix{
		\Par^{\text{\rm int}} ( \bfn)  \ar@{->>}[rr]^{q_2} \ar@{_{(}->}[d]_{\iota_1} & & \Par(\bfnred) \ar[d]^{{\mathcal L}_\text{\rm red}} \\
		\Par(\bfn) \ar[rr]^{\mathcal L} & & \R
	}
	\]
	
	\item 	For any $\mathbf{T} \in \Par^{\text{\rm int}} ( \bfn) $, we have:
	$ q_1 \left( \nabla_{\iota_1(\mathbf{T})}  {\mathcal L} \right) = \iota_2 \left( \nabla_{q_2(\mathbf{T})} {\mathcal L}_\text{\rm red}  \right). $
\end{enumerate}
\end{lemma}



\begin{proof}
We have the following standard inclusions into the first coordinates and projections onto the first coordinates, for  $\inx = 0, 1, \dots, L$:
\[ {\inc}_\inx  = \inc_{\nred_\inx, n_\inx} : \R^{\nred_{\inx}} \hookrightarrow \R^{n_\inx}, \qquad  \widetilde{\inc}_\inx = \inc_{1+ \nred_\inx, 1+ n_\inx} : \R^{1+ \nred_{\inx}} \hookrightarrow \R^{1+ n_\inx}, \]
\[ \pi_\inx : \R^{n_\inx} \rightarrow \R^{\nred_\inx}, \qquad \qquad \widetilde{\pi}_\inx :  \R^{1+ n_\inx} \rightarrow \R^{1+ \nred_\inx}. \]
Observe that  $\Par^{\text{\rm int}} ( \bfn) $ is the subspace of $\Par(\bfn)$ consisting of those $\mathbf{T} = (T_1, \dots, T_L) \in \Par(\bfn)$ such that:
\[ \left(\id_{n_\inx} - \inc_\inx \circ \pi_\inx\right) \circ T_\inx \circ \widetilde{\inc}_{\inx -1} \circ \widetilde{\pi}_{\inx -1} = 0 \]
for $\inx = 1, \dots, L$.  


By the definition of radial rescaling functions, for each $i = 1, \dots, L$, there is a piece-wise differentiable function $h_i : \R \to \R$ such that $\rho_i = h_i^{(n_i)}$. Note that $\rhored_i = h_i^{(\nred_i)}$, and $h^{(n_i)}  \circ \inc_i = \inc_i \circ h^{(\nred_i)}$.

The identity $\iota = \iota_1 \circ \iota_2$ follows directly from definitions. To prove the commutativity of the first diagram, it is enough to show that, for any $\bfX$ in $\Par(\bfnred)$, the feedforward  functions  of $\bfX$ and $\iota(\bfX)$ coincide. This follows easily from the fact that,  for $\inx = 1, \dots, L$, we have:
\[ {\pi}_\inx \circ h^{(n_\inx)} \circ \inc_{\inx} ={\pi}_\inx  \circ \inc_{\inx}\circ   h^{(\nred_\inx)} = h^{(\nred_\inx)}. \]
For the second claim, let $\mathbf{T}\in \Par^{\text{\rm int}} ( \bfn) $.  It suffices to show that $\iota_1(\bfT)$ and $q_2(\bfT)$ have the same feedforward function. Recall the $\ext_i$ maps and the formulation of the feedforward function in the merged notation given in Equation \ref{eq:feedforward-merged}. Using this set-up, the key computation is:
\begin{align*}
	\inc_\inx \circ h^{(\nred_\inx)} \circ \pi_\inx \circ  {T}_\inx \circ \ext_{n_{\inx -1}} \circ {\inc}_{\inx -1}  
	&=	h^{(n_\inx)}  \circ \inc_\inx \circ  \pi_\inx \circ {T}_\inx \circ \widetilde{\inc}_{\inx -1} \circ \ext_{n_{\inx -1}}    \\
	&=	h^{(n_\inx)} \circ {T}_\inx \circ \widetilde{\inc}_{\inx -1} \circ \ext_{n_{\inx -1}}    \\
	&=	h^{(n_\inx)} \circ {T}_\inx \circ \ext_{n_{\inx -1}}  \circ {\inc}_{\inx -1}   
\end{align*}
which uses the fact that $\left(\id_{n_\inx} - \inc_\inx \circ \pi_\inx\right) \circ T_\inx \circ \widetilde{\inc}_{\inx -1} = 0$, or, equivalently, $  \inc_\inx \circ \pi_\inx \circ T_\inx \circ \widetilde{\inc}_{\inx -1} = T_\inx \circ \widetilde{\inc}_{\inx -1}$, as well as the fact that $\ext_{i} \circ \inc_{i} = \widetilde{\inc}_i \circ \ext_i$. Applying this relation successively starting with the second-to-last layer $(\inx = L-1)$ and ending in the first $(\inx=1)$, one obtains the result. 
%
For the last claim, one computes $\nabla_{\mathbf{T}}( {\mathcal L} \circ \iota_1)$ in two different ways. The first way is: 
\begin{align*}
	\nabla_{\mathbf{T}}( {\mathcal L} \circ \iota_1) & = \left(   d ( {\mathcal L}_\bfT \circ \iota_1 )   \right)^T =  \left(   d {\mathcal L}_{\iota_1(\bfT)}  \circ d_\bfT \iota_1    \right)^T  =   \left(   d  {\mathcal L}_{\iota_1(\bfT)}   \circ \iota_1 \right)^T \\
	&= \iota_1^T \left(     d  {\mathcal L}_{\iota_1(\bfT)}^T \right) =  q_1 \left(  \nabla_{\iota_1(\bfT)}  {\mathcal L}  \right) 
\end{align*}
where we use the fact that $\iota_1$ is a linear map whose transpose is $q_1$. The second way uses the commutative diagram of the second part of the Lemma: 
\begin{align*}
	\nabla_{\mathbf{T}}( {\mathcal L} \circ \iota_1) & = \nabla_\bfT \left( {\mathcal L}_\text{\rm red} \circ q_2  \right) = \left(   d  \left( {\mathcal L}_\text{\rm red} \right)_\bfT \circ q_2    \right)^T =  \left(   d \left(  \mathcal L_\text{\rm red} \right)_{q_2(\bfT)}  \circ d \left(q_2  \right)_\bfZ   \right)^T \\ & =   \left(   d \left( {\mathcal L}_\text{\rm red} \right)_{q_2(\bfT)}  \circ q_2 \right)^T 
	= q_2^T \left(    d \left(  {\mathcal L}_\text{\rm red}  \right)_{q_2(\bfT)}^T \right) =  \iota_2 \left(  \nabla_{q_2(\bfT)}  {\mathcal L}_\text{\rm red}   \right).
\end{align*}
We also use the fact that $q_2$ is a linear map whose transpose is $\iota_2$. 
\end{proof}



\begin{proof}[Proof of Theorem \ref{thm:proj-gd}]  
As above, let $\bfR, \bfQ = \texttt{QR-compress}(\bfA)$ be the outputs of Algorithm \ref{alg:QR-mod-comp}, so that $\bfV = (\bfWred, \bfbred)\in \Par(\bfnred)$ is the dimensional reduction of the merged parameters $\bfA= (\bfW, \bfb)$, and $\bfQ\in O(\bfn^\text{\rm hid})$. Set $\bfT = \bfQ\inv \cdot \bfA  \in \Parintn$. 

The action of $\bfQ \in O(\bfn^\text{\rm hid})$ on $\Parn$ is an orthogonal transformation, so the first claim follows from Lemma \ref{lem:orth-gd}.

For the second claim,  it suffices to consider the case $\eta = 1$. The general case follows similarly.  We proceed by induction. The base case $k=0$ amounts to Theorem~\ref{thm:mod-comp}. For the induction step, we set
\[ \mathbf{Z}^{(k)}  = \iota(	\gamma_\text{\rm red}^k (\bfV) ) + \bfT - \iota( \bfV). \]
Each $\mathbf{Z}^{(k)}$ belongs to $\Par^{\text{\rm int}} ( \bfn) $, so  $i_1( \mathbf{Z}^{(k)} ) = \mathbf{Z}^{(k)}$. Moreover,  $q_2 \left( \mathbf{Z}^{(k)}  \right) = \gamma_\text{\rm red}^k (\bfV) $.
We compute:
\begin{align*}
	\gamma_\text{\rm proj}^{k+1}( \bfQ\inv \cdot \bfA) & = \gamma_\text{\rm proj}\left(   \gamma_\text{\rm proj}^k(\bfQ\inv \cdot \bfA) \right)  \\ 
	&=   \gamma_\text{\rm proj} \left(  \iota(  \gamma_\text{\rm red}^k(\bfV)) + \bfT - \iota( \bfV) \right) \\
	&=     \iota_1 \circ q_1  \left ( \iota(  \gamma_\text{\rm red}^k(\bfV)) +    \bfT - \iota( \bfV)  -  \nabla_{ \iota(  \gamma_\text{\rm red}^k(\bfV)) + \bfT - \iota( \bfV) } {\mathcal L}    \right)\\
	&= \iota(  \gamma_\text{\rm red}^k(\bfV))  -    \iota_1 \circ q_1 \left( \nabla_{ \iota_1(\mathbf{Z}^{(k)}) } {\mathcal L}     \right)  +  \bfT - \iota( \bfV) \\
	&= \iota(  \gamma_\text{\rm red}^k(\bfV)) -    \iota_1 \circ \iota_2  \left( \nabla_{ q_2(\mathbf{Z}^{(k)}) } {\mathcal L}_\text{\rm red}  \right) +  \bfT - \iota( \bfV) \\
	&=  \iota \left( \gamma_\text{\rm red}^k(\bfV) -  \nabla_{  \gamma_\text{\rm red}^k (\bfV)  }{\mathcal L}_\text{\rm red} \right ) +   \bfT - \iota( \bfV) \\
	&= \iota \left(  \gamma_\text{\rm red}^{k+1}(\bfV)   \right) +   \bfT - \iota( \bfV)
\end{align*}
where the second equality uses the induction hypothesis; the third invokes the definition of ${\gamma}_\text{\rm proj}$; the fourth  uses  the fact that $\mathbf{Z}^{(k)}=\iota(	\gamma_\text{\rm red}^k (\bfV) ) + \bfT - \iota( \bfV)$ belongs to $\Parintn$; the fifth and sixth  use  Lemma \ref{lem:repint} above; and the last uses the definition of $ \gamma_\text{\rm red}$. 	
\end{proof}

\subsection{Example}  \label{appsubsec:ex-131}

We now discuss an example where projected gradient descent does not match usual gradient descent. 

Let $\bfn = (1, 3,1)$ be a widths vector. The space of parameters  with this widths vector is 10-dimensional: \[ \Par( \bfn) = \Hom(\R^2, \R^3) \oplus \Hom(\R^4, \R) \simeq \R^{10}. \]
We identify  a choice of parameters (in the merged notation)
\begin{equation}\label{eq:ex-parameters}
\bfA = \left( \ A_1 = \begin{bmatrix}
	a &  b \\ c & d \\ e & f 
\end{bmatrix} \ , \  A_2  =  \begin{bmatrix}
	g & h  & i & j
\end{bmatrix} \right) \in  \Par((1,3,1))
\end{equation}
with the point $p = (a,b,c,d, e,f, g, h, i , j)$ in $\R^{10}$. To be even more explicit, the weights for the first layer are $W_1 = \begin{bmatrix}
b \\ d \\ f
\end{bmatrix} $, the bias in the first hidden hidden layer is $b_1 = (a, c, e)$, the weights for the second layer are $W_2 = \begin{bmatrix}
h & i & j
\end{bmatrix}$, and the bias for the output layer is $b_2 = g$.

The action of the orthogonal group $O(\bfn) = O(3)$ on $\Par( \bfn) \simeq \R^{10}$ can be expressed as:
\[ Q \mapsto \begin{bmatrix}
Q & 0 & 0 & 0 \\
0 & Q & 0 & 0 \\
0 & 0 & 1 & 0 \\
0 & 0 & 0 & Q
\end{bmatrix},
\]
where the rows and columns are divided according to the partition $3 + 3 +1+ 3=10$. 
Consider the function\footnote{
For $\bfA \in \Par( \bfn)$, the neural function of the neural network with affine maps determined by $\bfA$ and identity activation functions is $\R \to \R$;  $x \mapsto \mathcal L(\bfW)x$.  
The function $\mathcal L$ can appear as a loss function for certain batches of training data and cost function on $\R$.}:
\begin{align*} \mathcal L : \Par( \bfn)  &\to \R \\
p = (a,b,c,d, e, f, g, h, i, j) &\mapsto h(a+b) + i (c + d) + j (e + f) + g
\end{align*}
By the product rule, we have:
\[ \nabla_{p} {\mathcal L} = (h, h, i , i , j , j , 1 , a +b, c + d, e + f) \]
One easily checks that $\mathcal L ( Q \cdot p) = \mathcal L(p)$ and that $\nabla_{Q \cdot p } {\mathcal L} = Q \cdot \nabla_{p} {\mathcal L}$ for any $Q \in O(3)$.

The interpolating space is the eight-dimensional subspace of $\Par( \bfn) \simeq \R^{10}$ with $e = f = 0$ (using the notation of Equation \ref{eq:ex-parameters}). Suppose $p^\prime = (a, b, c,d, 0 , 0, g, h, i, j)$ belongs to the interpolating space. Then the gradient is 
\[ \nabla_{p^\prime} \mathcal L = (h, h, i , i , j ,j , 1, a+ b, c+ d, 0) \]
which does not belong to the interpolating space.  So one step of usual gradient descent, with learning rate $\eta >0$ yields:
\begin{align*} \gamma : & p^\prime =    (a,b,c,d, 0, 0, g, h , i , j) \mapsto  \\
&(a-\eta h \ , \ b - \eta h \ , \ c - \eta i \ , \ d - \eta i \ , \ -\eta j  \ , \ -\eta j  \ , \  g - \eta  \ , \ h - \eta (a+ b) \ , \  i - \eta(c+d) \ , \ j  )
\end{align*}
On the other hand, one step of projected gradient descent yields:
\begin{align*}
\gamma_{\text{\rm proj}}   : p^\prime   &= (a,b,c,d, 0, 0, g, h , i , j) \mapsto \\
&(a-\eta h \ , \ b - \eta h \ , \ c - \eta i \ , \ d - \eta i \ , \  0  \ , \ 0  \ , \  g - \eta  \ , \ h - \eta (a+ b) \ , \  i - \eta(c+d) \ , \ j  )
\end{align*}
Direct computation shows that the difference between the evaluation of $\mathcal L$ after one step of gradient descent and the evaluation of $\mathcal L$ after one step of projected gradient descent is:
\[ \mathcal L( \gamma( p^\prime) )  - \mathcal L(\gamma_{\text{\rm proj}}(p^\prime)) = 2 \eta j^2. \]

%

\section{Experiments}\label{app:exp}

As mentioned in Section \ref{sec:exp}, we provide an implementation of  
Algorithm \ref{alg:QR-mod-comp} in order to 
(1) empirically validate that our implementation satisfies the claims of Theorems \ref{thm:mod-comp} and Theorem \ref{thm:proj-gd} and (2) quantify real-world performance. Our implementation uses a generalization of radial neural networks, which we explain presently. 

\subsection{Radial neural networks with shifts}\label{appsubsec:withshifts}

In this section, we consider radial neural networks with an extra trainable parameter in each layer that shifts the radial rescaling activation. Adding such parameters allows for more flexibility in the model, and (as shown in Theorem \ref{thm:withshift}) the model compression of Theorem \ref{thm:mod-comp} holds for such  networks. It is this generalization that we use in our experiments.

Let $h : \R \to \R$ be a function. For any $n \geq 1$ and any $t \in \R$, the corresponding {\it shifted radial rescaling function} on $\R^n$ is  given by:
\[ \rho =  h^{(n,t)}: v \mapsto \frac{h(\lvert v \rvert - t)}{|v|} v\]
if $v \neq 0$ and $\rho(0) = 0$.
A {\it radial neural network with shifts}  consists of the following data:  
\begin{enumerate}
\item Hyperparameters: A positive integer $L$ and a  widths  vector $\mathbf{n} = (n_0, n_1, n_2, \dots, n_L)$.
\item Trainable parameters:
\begin{enumerate}
	\item A choice of weights and biases $(\bfW, \bfb) \in \Par(\bfn).$
	\item A vector of shifts  $\mathbf{t} = (t_1, t_2, \dots, t_L)  \in \R^L$. 
\end{enumerate}

\item Activations:  A tuple $\bfh = (h_1, \dots, h_L)$ of piecewise differentiable functions $\R \to \R$. Together with the shifts, we have the shifted radial rescaling activation $\rho_i = h_i^{(n_i, t_i)} : \R^{n_i} \to \R^{n_i}$ in each layer. 
\end{enumerate}



The {\it  feedforward function} of a radial neural network with shifts is defined in the usual recursive way, as in Section \ref{sec:rad-nns}. 
The trainable parameters form the vector space $\Par(\bfn) \times \R^L$, and the loss function  of a batch of training data $\{(x_i, y_i)\} \subset \R^{n_0} \times \R^{n_L}$ is defined as
\begin{align*} \L :  \Par( \mathbf{n}) \times \R^L  \longrightarrow \R ; \qquad \qquad 
(\mathbf{W} , \mathbf{t}) \mapsto \sum_j \mathcal C(F_{(\mathbf{W}, \bfb,  \mathbf{t},  \bfh )} ( x_j),  y_j)
\end{align*}
where $F_{(\bfW, \bfb,\bft, \bfh)}$ is the feedforward function of a radial neural network with weights $\bfW$, biases $\bfb$, shifts $\bft$, and radial rescaling activations produced from $\bfh$.
We have the  gradient descent map:
\begin{align*} \gamma :  \Par( \mathbf{n}) \times \R^L  &\longrightarrow \Par( \mathbf{n}) \times \R^L 
\end{align*}
which updates the entries of $\bfW$, $\bfb$, and  $\mathbf{t}$. The group $O(\bfn^\text{\rm hid}) = O(n_1) \times \cdots \times O(n_{L-1})$ acts on $\Par( \bfn)$ as usual (see Section \ref{subsec:paramspace}), and on $\R^L$ trivially. The neural function is unchanged by this action. We conclude that  the $O(\bfn^\text{\rm hid})$ action on $\Par( \bfn) \times \R^L$ commutes with gradient descent $\gamma$. 
We now state a generalization of Theorem \ref{thm:mod-comp} for the case of radial neural networks with shifts. We omit a  proof, as it uses the same techniques as the proof of Theorem \ref{thm:mod-comp}.

\begin{theorem}\label{thm:withshift} 
Let $(\bfW, \bfb, \bft, \bfh  )$ be a radial neural network with shifts and widths vector $\bfn$. Let $\bfWred$ and $\bfbred$ be the  weights and biases of the compressed network produced by Algorithm \ref{alg:QR-mod-comp}.
The feedforward function of the original network  $(\bfW, \bfb, \bft, \bfh  )$ coincides with that of the compressed network   $(\bfWred, \bfbred, \bft, \bfh  )$.

\end{theorem}

Theorem \ref{thm:proj-gd} also generalizes to the setting of radial neural networks with shifts, using projected gradient descent with respect to the subspace  $\Parintn \times \R^{L}$ of $\Parn \times \R^L$.

\subsection{Implementation details}\label{subsec:implementation}

Our implementation is written in Python and uses the QR decomposition routine in NumPy \cite{harris2020array}.  We also implement a general class \texttt{RadNet} for radial neural networks using PyTorch \cite{NEURIPS2019_9015}.
For brevity, we write $\hbfW$ for $(\bfW, \bfb)$ and $\hbfWred$ for $(\bfWred, \bfbred)$.

\paragraph{(1)  Empirical verification of Theorem \ref{thm:mod-comp}.}\label{subsec:exp-QR-thm}
We use synthetic data to learn the function $f(x) = e^{-x^2}$ with $N=121$ samples $x_j = -3 + j/20$ for $0 \leq j < 121$.  We model $f_{\hat{\bfW}}$ as a radial neural network with widths $\bfn = (1,6,7,1)$ and activation the radial shifted sigmoid $h(x) = 1/(1+e^{-x + s})$.  Applying \texttt{QR-compress} gives a radial neural network $f_{\hat{\bfW}^\text{\rm red}}$ with widths $\bfn^{\mathrm{red}} = (1,2,3,1)$.  Theorem \ref{thm:mod-comp} implies that the neural functions of  $f_{\hat{\bfW}}$ and $f_{\hat{\bfW}^\text{\rm red}}$ are equal.  Over 10 random initializations of $\hat{\bfW}$, the mean absolute error $(1/N) \sum_{j} |f_{\hat{\bfW}}(x_j) - f_{\hat{\bfW}^\text{\rm red}}(x_j)| = 1.31 \cdot 10^{-8} \pm 4.45 \cdot 10^{-9}$. Thus $f_{\hat{\bfW}}$ and $f_{\hat{\bfW}^\text{\rm red}}$ agree up to machine precision.  

\paragraph{(2) Empirical verification of Theorem \ref{thm:proj-gd}.}\label{subsec:exp-projGD-thm}
Adopting the notation from above, the claim  is that training $f_{\bfQ\inv \cdot \hbfW}$ with objective $\mathcal{L}$ by projected gradient descent coincides with training  $f_{\hbfWred}$ with objective $\mathcal{L}_{\mathrm{red}}$ by usual gradient descent. We verified this on synthetic data using 3000 epochs at learning rate 0.01. Over 10 random initializations of $\hbfW$, the loss functions match up to machine precision with $|\mathcal{L}-\mathcal{L}_{\mathrm{red}}| = 4.02 \cdot 10^{-9} \pm 7.01 \cdot 10^{-9}$.

\paragraph{(3) Reduced model trains faster.}  \label{subsec:exp-faster}
Due to the relation between projected gradient descent of the full network $\hbfW$ and gradient descent of the reduced network $\hbfWred$, our method may be applied before training to produce a smaller model class which \emph{trains} faster without sacrificing accuracy. We test this hypothesis in learning the function $f : \R^2 \to \R^2$ sending $x= (t_1, t_2)$ to $(e^{-t_1^2}, e^{-t_2^2})$ using $N=121^2$ samples $( -3 + j/20, -3 + k/20)$ for $0 \leq j, k < 121$. We model $f_{\hbfW}$ as a radial neural network with layer widths $\bfn = (2,16,64, 128, 16, 2)$ and activation the radial  sigmoid $h(r) = 1/(1+e^{-r})$.  Applying  \texttt{QR-compress}  gives a radial neural network $f_{\hbfWred}$ with widths $\bfn^{\mathrm{red}} = (2,3,4,5,6,2)$. We trained both models until the training loss was $\leq0.01$.  Running on a system with an Intel i5-8257U@1.40GHz and 8GB of RAM and averaged over 10 random initializations, the reduced network trained in $15.32 \pm 2.53$ seconds and the original network trained in $31.24 \pm 4.55$ seconds.

\paragraph{(4) Comparison with ReLU MLP on noisy image recovery.} We show that a Step-ReLU radial network performs better than an otherwise comparable network with pointwise ReLU on a noisy image recovery task. Using samples of MNIST with significant added noise the network classification task is to identify from which original sample the noisy sample derives. 

Specifically, we choose \texttt{n} samples from MNIST, all with the same MNIST label, and produce \texttt{m} noisy samples from each by adding noise. The noise is added by considering each sample as a point in $\R^{784}$, and adding uniform random noise in a ball around each. The radius of the ball around a given point is the product of the noise level variable (\texttt{noise\_scale}, which is the same for all points) and the minimal distance to another sample point (which varies from point to point). As indicated in Figure \ref{fig:noisy-threes}, when \texttt{noise\_scale}=3 the classification task is difficult for the human eye. 

Our data takes \texttt{n} = 3 original MNIST images with the same label, and produces \texttt{m} = 100 noisy images for each, with \texttt{noise\_scale}=3. We perform a 240 train / 60 test split of the 300 data points. Both models have three layers with widths $(d, d+1, d+2, \mathtt{n} =3)$, where $d = 28^2 = 784$; hence, both models have $620,158$ trainable parameters

Over 10 trials, each training for 150 epochs and learning rate 0.05 for both models, the radial network achieves training loss 0.00256 $\pm 3.074 \cdot 10^{-4}$ with accuracy 1 $\pm$ 0, while the ReLU MLP has training loss 0.00393 $\pm3.992 \cdot 10^{-4}$ with accuracy 1 $\pm$ 0. On the test set, the radial network has loss 0.00266 $\pm 3.749 \cdot 10^{-4}$ with accuracy 1 $\pm$ 0, while the  ReLU MLP has loss 0.0041 $\pm 4.442 \cdot 10^{-4}$ with accuracy 1 $\pm$ 0.
The convergence rates are illustrated in Figure \ref{fig:noisy-threes}, with the radial network outperforming the ReLU MLP. We note that 150 epochs is sufficient for all methods to converge.

We observe that the radial network 1) is able to obtain a better fit, 2) has faster convergence, and 3) generalizes better than the pointwise ReLU.  We hypothesize the radial nature of the random noise makes radials networks well-adapted to the task.


\section{Relation to  radial basis function networks}\label{app:rbfns}

In this appendix, we show that radial neural networks are equivalent to a particular class of multilayer radial basis functions networks. This class is obtained by imposing the condition that the so-called `hidden dimension' at each layer is equal to one; the total number of layers, however, is unconstrained. To our knowledge, the literature contains no universal approximation result for this class of radial basis functions networks.


\subsection{Single layer case}

\newcommand{\rbfn}{RBFN }

We first recall the definition of a  radial basis function network. 
A {\it local linear model extension of a radial basis function network} (henceforth abbreviated simply by {\it RBFN})  consists of:
\begin{itemize}
	\item An input dimension $n$, an output dimension $m$, and a `hidden' dimension $N$.
	\item For $i = 1, \dots, N$, a matrix $W_i \in \R^{m \times n}$, a vector $b_i \in \R^n$, and a weight $a_i \in \R^m$.
	\item A nonlinear function\footnote{A more general version allows for a different nonlinear function for every $i = 1, \dots, N$.} $\lambda : \R \to \R$. 
\end{itemize}
The feedforward function of a \rbfn  is defined as:
$$F : \R^n \to \R^m \qquad \qquad x \mapsto \sum_{i=1}^N \left(a_i + W_i (x + b_i) \right)\lambda(|x + b_i|) .$$
The integer $N$ is commonly referred to as `the hidden number of neurons'. This is a bit of a misnomer. Really there is only one layer with input dimension $n$ and output dimension $m$; the integer $N$ is part of the specification of the activation function. 

We observe that if $N=1$ and $a_1 =0$, then the feedforward function  is given by:
$$F : \R^n \to \R^m \qquad \qquad x \mapsto W \rho(x+b)$$
where $\rho$ is the radial rescaling function determined by $\lambda$. 
In words, one adds  $b_1 = b \in \R^n$ to the input vector $x$, applies the activation $\rho$ to obtain new vector in $\R^n$, and then applies the linear transformation determined by the matrix $W_1 = W$ to obtain the output vector in $\R^m$. Motivated by this observation, we say that a RBFN is {\it constrained} if $N=1$ and $a_1 =0$.

\subsection{Constrained multilayer case}

Next, we consider the constrained multilayer case of a radial basis functions network. Specifically, a {\it constrained multilayer} RBFN consists of:
\begin{itemize}
	\item A widths vector $(n_0, \dots, n_L)$ where $L$ is the number of layers.
	\item A matrix $W_\ell \in \R^{n_\ell \times n_{\ell -1}}$ for $\ell = 1, \dots, L$.
	\item A vector $b_{\ell} \in \R^{n_{\ell}}$ for $\ell = 0, 1, \dots, L-1$.
	\item A nonlinear function $\lambda_\ell : \R \to \R$ for $\ell =0, 1, \dots, L-1$. (Equivalently, the corresponding radial rescaling function $\rho_\ell : \R^{n_{\ell}} \to \R^{n_{\ell}}$ for  $\ell = 0, \dots, L-1$.)
\end{itemize}
The feedforward function is defined as follows. For $\ell = 0, \dots, L$, we recursively define $F_\ell : \R^{n_0} \to \R^{n_\ell}$ by setting $F_0(x) = x$ and 
$$F_\ell(x) = W_\ell \rho_{\ell-1}(F_{\ell-1}(x)+b_{\ell-1})$$
for $\ell = 1, \dots, L$. The feedforward function is $F_L$. 


\subsection{Relation to radial neural networks}

We now demonstrate that radial neural networks are equivalent to constrained multilayer RBFNs.

\begin{proposition}
	For any radial neural network, there is a constrained multilayer RBFN with the same feedforward function. Conversely, for any constrained multiplayer RBFN, there is a radial neural network with the same feedforward function.
\end{proposition}

\begin{proof} For the first statement, let $(\bfW, \bfb, \boldrho)$ be a radial neural network with $L$ layers and widths vector $(n_0, \dots, n_L)$. Recall the partial feedforward functions $G_\ell : \R^{n_0} \to \R^{n_\ell}$  defined recursively by setting $G_0(x) = x$ and
	$$G_\ell(x) = \rho_\ell \left(W_\ell G_{\ell -1}(x) + b_\ell \right)$$
	The feedforward function is $G_L$. 
	Consider the constrained multilayer RBFN with $L+1$ layers and the following:
	\begin{itemize}
		\item Widths vector $(n_0, n_1,\dots, n_{L-1},  n_L, n_L)$. The last two layers have the same dimension.
		\item Weight matrices $W_\ell \in \R^{n_\ell \times n_{\ell -1}}$ for $\ell = 1, \dots, L$ and $W_{L+1} = \id_{n_L} \in \R^{n_L \times n_L}$.
		\item A vector $b_{\ell} \in \R^{n_{\ell}}$ for $\ell = 1, \dots, L$, and $b_0 = 0 \in \R^{n_0}$.
		\item A radial rescaling activation ${\rho}_\ell : \R^{n_\ell} \to \R^{n_\ell}$ for  $\ell = 1, \dots, L$, and $\rho_0 = \id_{n_0}$. 
	\end{itemize}
	Let $F_\ell$ be the partial feedforward functions for this RBFN, defined recursively as above. We claim that $$F_\ell (x) = W_\ell \circ G_{\ell -1}(x)$$ for any $x \in \R^{n_0}$ and $\ell = 1, \dots, L$. We prove this by induction. The base case is $\ell = 1$:
	$$F_1(x) = W_1 \circ \rho_0 \left( F_0(x) + b_0 \right) = W_1 x = W_1 \circ G_0(x)$$
	For the induction step, take $\ell >1$ and compute:
	$$F_\ell(x) = W_\ell \circ \rho_{\ell - 1} \left( F_{\ell -1}(x) + b_{\ell -1} \right) =  W_\ell \circ \rho_{\ell - 1} \left( W_{\ell -1}G_{\ell -2}(x) + b_{\ell -1} \right)  = W_\ell  \circ G_{\ell -1}(x)  $$
	The first claim now follows from the case $\ell = L$, using the fact that $W_{L+1}$ is the identity. 
	
	For the second statement, let  $(\bfW, \bfb, \boldrho)$ be a constrained multilayer RBFN with $L$ layers and widths vector $(n_0, \dots, n_L)$. Consider the radial neural network with $L+1$ layers and the following:
	\begin{itemize}
		\item Widths vector $(n_0, n_0, n_1,\dots, n_{L-1},  n_L)$. The first two layers have the same dimension.
		\item Weight matrices given by $\tilde{W}_1 = \id_{n_0}$ and $\tilde{W}_\ell = W_{\ell-1}$ for  $\ell = 2, \dots, L+1$.  
		\item Bias vectors given by $\tilde{b}_{\ell} = b_{\ell-1}$  for $\ell = 1, 2, \dots, L$, and $\tilde{b}_{L+1} = 0$. 
		\item Radial rescaling activations given by $\tilde{\rho}_\ell  = {\rho}_{\ell-1}$ for  $\ell = 1, \dots, L$, and $\tilde{\rho}_{L+1} = \id_{n_L}$. 
	\end{itemize}
	One uses the recursive definition of the partial feedforward functions to show that, for $\ell = 1, \dots, L$, we have $F_\ell(x) = W_\ell \circ G_\ell(x)$, where $F_\ell$ and $G_\ell$ are the partial feedforward functions of the RBFN and radial neural network, respectively. Then:
	$$G_{L+1} (x) = \tilde{\rho}_{L+1} \left(\tilde{W}_{L+1} \circ G_L(x)  + \tilde{b}_{L+1} \right) = W_L \circ G_L(x) = F_L(x),$$
	so the two feedforward functions coincide.
\end{proof}

\subsection{Conclusions}

While radial neural networks are equivalent to a certain class of radial basis function network, we point out differences between our results  and the standard theory of radial basis functions network. First,  RBFNs generally only have two layers; we consider ones with unbounded depth. Second, to our knowledge, ours is the first universal approximation result such that:
\begin{itemize}
	\item  it uses networks in the subclass of  multilayer RBFNs satisfying  the constraint that all the number of `hidden neurons' in each layer is equal  to $1$. 
	\item it approximates functions with networks of bounded width.	
	\item it can be used to approximate asymptotically affine functions, rather than functions defined on a compact domain. 
\end{itemize}
Our compressibility result  may apply to multilayer RBFNs where the number of `hidden neurons' $N_\ell$ at each layer is not equal to $1$, but we expect the compression to be weaker, and that constrained mulitlayer RBFNs are in some sense the most compressible type of RBFN.

\end{document}